\documentclass[10pt,twocolumn,letterpaper]{article}

\usepackage{iccv}              

\usepackage{pifont}
\usepackage{xspace}
 
\newcommand{\cmark}{\ding{51}}%
\newcommand{\xmark}{\ding{55}}
\newcommand{\yes}{\textcolor{ForestGreen}{\cmark}}
\newcommand{\no}{\textcolor{Bittersweet}{\xmark}}

\DeclareMathOperator{\logsumexp}{LogSumExp}
\DeclareMathOperator{\softmax}{Softmax}
\newcommand{\POPE}[1]{POPE~\cite{pope}\xspace}
\newcommand{\MMVP}[1]{MMVP~\cite{mmvp}\xspace  }
\newcommand{\MME}[1]{MME~\cite{mme}\xspace  }
\newcommand{\VCD}[1]{VCD~\cite{vcd}\xspace  }    
\newcommand{\ICD}[1]{ICD~\cite{wang2024ICD}\xspace}
\newcommand{\HALC}[1]{HALC~\cite{halc}\xspace }
\newcommand{\PAI}[1]{PAI~\cite{pai}\xspace }
\newcommand{\OPERA}[1]{OPERA~\cite{huang2023opera}\xspace}
\newcommand{\DOLA}[1]{DoLa~\cite{dola}\xspace}
\newcommand{\CGD}[1]{CGD~\cite{deng2024seeing}\xspace}

\newcommand{\llavaimproved}[1]{LLaVA-1.5~\cite{llava_improved}\xspace}
\newcommand{\instructblip}[1]{InstructBLIP~\cite{instructblip}\xspace}
\newcommand{\mplugowl}[1]{mPLUG-Owl2~\cite{mplugowl2}\xspace}
\newcommand{\best}[1]{{$\textbf{#1}$}}
\newcommand{\second}[1]{{$\underline{#1}$}}

\definecolor{commentcolor}{RGB}{110,134,185}   
\newcommand{\PyComment}[1]{\small\textcolor{commentcolor}{\# #1}}  
\newcommand{\PyCode}[1]{\small\textcolor{black}{#1}} 

\usepackage{multirow}

\usepackage{multicol}
\usepackage{adjustbox}
\usepackage{booktabs,tabularx, colortbl}

\definecolor{iccvblue}{rgb}{0.21,0.49,0.74}
\usepackage[pagebackref,breaklinks,colorlinks,allcolors=iccvblue]{hyperref}

\usepackage[dvipsnames]{xcolor}
\usepackage[ruled,vlined]{algorithm2e}
 
\usepackage{amsmath} 
\usepackage{caption} 
\usepackage{tcolorbox} 
 
\definecolor{mygray}{gray}{0.9}

\title{Energy-Guided Decoding for Object Hallucination Mitigation}
 
\author{Xixi Liu$^1$ \quad
Ailin Deng$^2$ \quad
Christopher Zach$^1$  \\
$^1$Chalmers University of Technology\\
$^2$National University of Singapore\\ 
{\tt\small xixil@chalmers.se}}
 
\begin{document}
\maketitle
 \begin{abstract}
 Mitigating object hallucination in large vision-language models (LVLMs) is critical to their safe deployment. Existing methods either are restricted to specific decoding methods, or demand sophisticated modifications to visual inputs, or rely on knowledge from external models. In this work, we first reveal the phenomenon that VLMs exhibit significant imbalance in the ``Yes'' ratio ( \ie, the fraction of ``Yes'' answers among the total number of questions) across three different visual question answering (VQA) datasets. Furthermore, we propose an energy-based decoding method, which dynamically selects the hidden states from the layer with minimal energy score. It is simple yet effective in reducing the bias for the yes ratio while boosting performance across three benchmarks (POPE, MME, and MMVP). Our method consistently improves accuracy and F1 score on three VQA datasets across three commonly used VLMs over several baseline methods. The average accuracy improvement is $4.82\%$ compared to greedy decoding. Moreover, the average yes-ratio gap reduction is $8.81\%$, meaning the proposed method is less biased as shown in Figure~\ref{yes_ratio_overview}.
\end{abstract}

 \section{Introduction}
\label{sec:intro}

Large language models (LLMs) such as ChatGPT~\cite{chatgpt} have shown great capability spanning over a wide range of domains including but not limited to search and personalized recommendation, virtual assistants, fraud detection,
and coding assistance tools. Meanwhile, vision-language models (VLMs) such as GPT-4V(ision) can describe the real world \eg, to visually impaired people~\cite{gpt4v, openflamingo, palm-e,llama-adaptor}. 
However, all those models, also known as foundation models, suffer from the issue of hallucination. Hallucination in LLMs refers to the problem that either the output of LLMs is inconsistent with the source content in context, or the LLMs generate a response that is not grounded by the pre-training dataset~\cite{weng2024hallucination}. Not surprisingly, all VLMs are also affected by hallucinations, which in this context refer to \emph{VLMs occasionally generating responses that are not supported by the visual input}. A recent survey~\cite{bai2024hallucinationmultimodallargelanguage} categories the hallucinations in VLMs, in particular, object-related hallucinations into the following groups: 1) \emph{category}, where the VLM identifies incorrect or non-existing objects in the image; 2) \emph{attribute}, where wrong descriptions such as color and shape for the given visual input are generated; 3) \emph{relation}, where incorrect relationships or interactions between objects are reported. 
Existing benchmarks used to assess the extent of hallucination in VLMs including POPE~\cite{pope}, MME~\cite{mme} and MMVP~\cite{mmvp}, which cover all three mentioned types of hallucination, with detailed information summarized in Table~\ref{tab:hallucination}. Equally important is the need to preserve the capability for open-ended generation while mitigating hallucinations. Therefore, Caption Hallucination Assessment with Image Relevance (CHAIR)~\cite{objectHallucination} is a primary metric considered in this work.

\begin{figure}[t]
    \centering
     \includegraphics[width=0.49\linewidth]{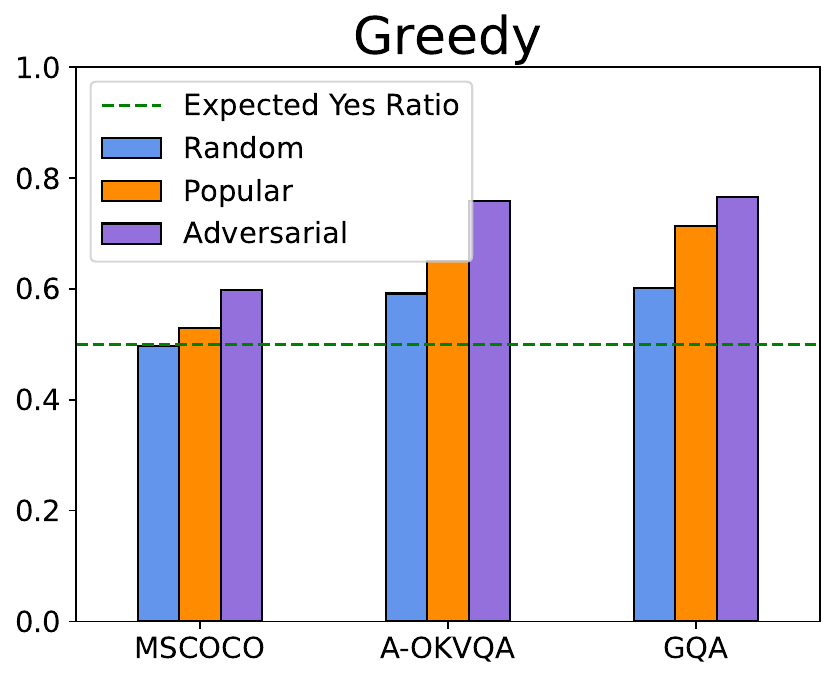}
      \includegraphics[width=0.49\linewidth]{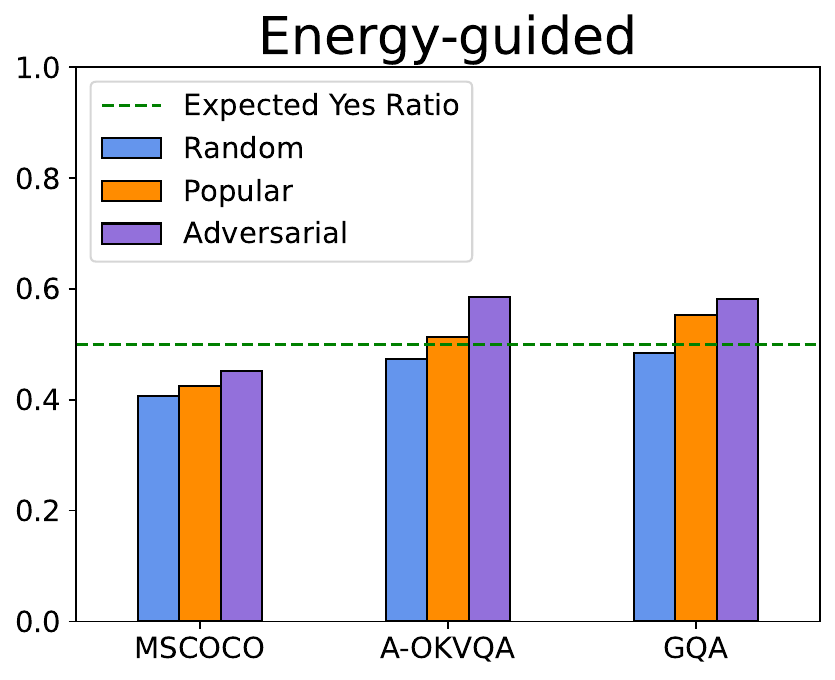}

    \caption{\emph{Yes Ratio using Greedy Decoding (left) and Energy-Guided Decoding (right)} across three datasets over three settings. LLaVA-1.5~\cite{llava_improved} is employed as the VLM backbone. The optimal yes ratio is $50\%$ (green dashed line). }

    \label{yes_ratio_overview}
\end{figure}

\begin{figure*}[!ht]
    \centering
    \includegraphics[width=0.7\linewidth]{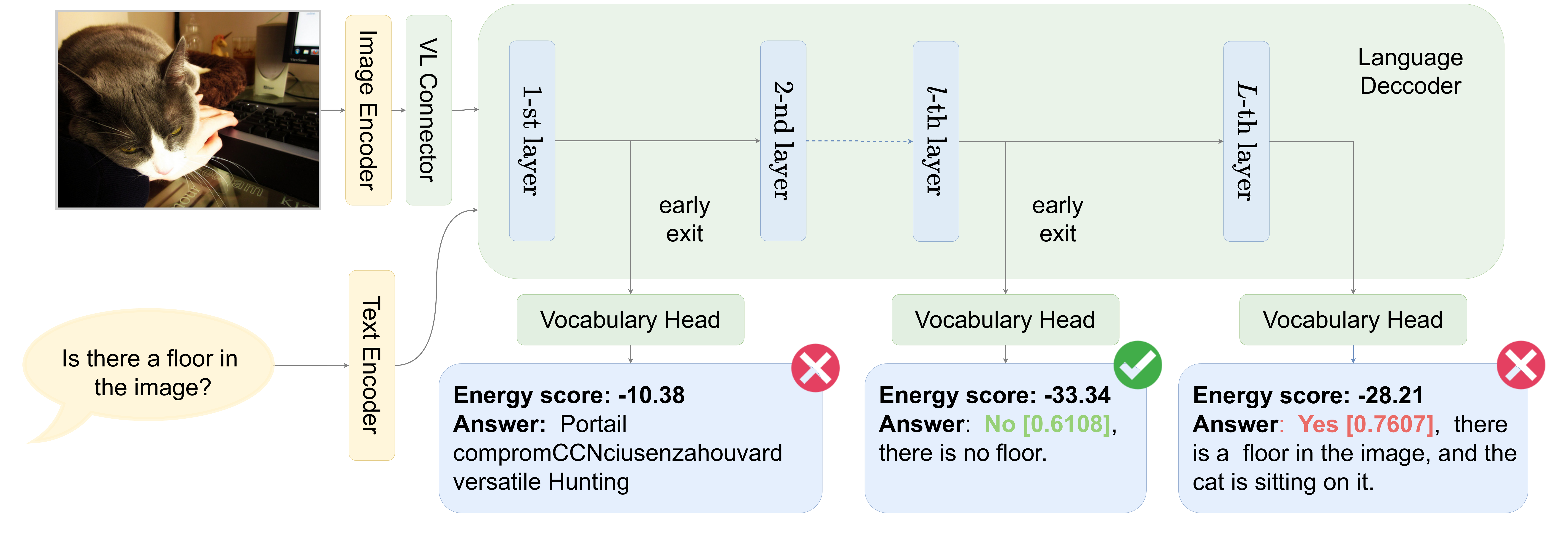}
      \includegraphics[width=0.28\linewidth]{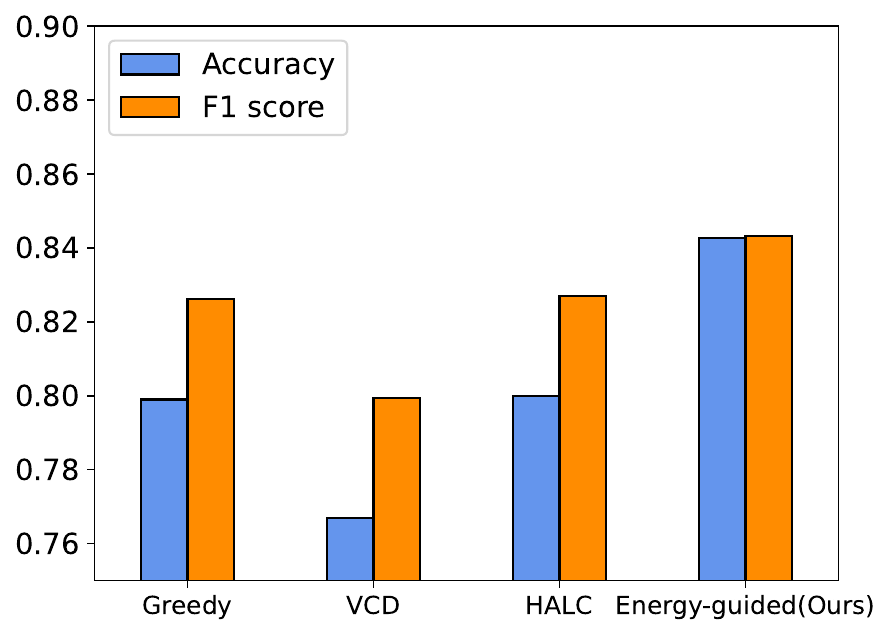}
    \caption{\emph{An Illustration of Energy-Guided Decoding.} We observe that the hidden states from the layer with minimal energy score generates more accurate responses. We also report the confidence of `` Yes'' and `` No'' measuring by the corresponding token probability. The right barplot shows the overall performance comparison in terms of accuracy and F1 score on GQA dataset with \emph{adversarial} setting and the VLM backbone is LLaVA-1.5~\cite{llava_improved}.  }
    \label{fig:toy_example}
\end{figure*}

 The problem of object hallucination mitigation can be traced back to~\cite{objectHallucination}, which is the initial work to investigate object hallucinations in the image captioning task. The cause of hallucination in VLMs is more complex. First, the hallucination might be induced by the language prior~\cite{vcd, huang2023opera}, which is analogous to hallucinations in LLMs~\cite{dola, wang2024ICD}. Second, a number of possible causes are related to the utilized visual encoders, such as its capacity~\citep{mmvp}, the quality of vision-language (VL) instruction-following data~\citep{lure}, insufficient attention to the image embedding during inference~\cite{pai}, and the training objectives employed for feature alignment~\cite{metamorph}.
 
Compared to prior works, our method does not require contrastive decoding~\cite{vcd, halc, wang2024ICD}, specific decoding strategies~\cite{deng2024seeing,huang2023opera}, corrupted images~\cite{vcd}, or prompt engineering~\cite{wang2024ICD}. It is highly efficient by only requiring a single forward pass to calculate the energy score at each layer. The hidden states from the layer with minimal energy score are then utilized for subsequent decoding.
 
\paragraph{Contributions}
\begin{enumerate}
    \item We empirically observe the inherent bias in terms of yes ratio that exists in the language decoder, particularly, for out-of-distribution datasets including Q-OKVQA~\cite{a-okvqa} and GQA~\cite{gqa}, cf.\ Figure~\ref{yes_ratio_overview} (left).
    \item Further, we propose a \emph{hyperparameter-free} decoding strategy termed ``energy-guided decoding.'' It does not require fine-tuning, contrastive decoding, or external models. Meanwhile, it performs very well, resulting in a less biased yes ratio, cf.\ Figure~\ref{yes_ratio_overview} (right) and improved accuracy and F1 score, cf.\ Figure~\ref{fig:toy_example} (right), while maintaining the capability of open-ended generation.
\end{enumerate}

 \section{Related work} 

\paragraph{Contrastive decoding in VLMs} 
 Contrastive decoding was initially proposed to mitigate hallucinations in LLMs~\cite{li2023contrastivedecodingopenendedtext}. Specifically, it leverages two LLMs with different capabilities (\ie, one is the ``expert'' and the other is the ``amateur''). By contrasting the predictive distribution from two LLMs, the token that captures the largest difference is selected for generation. DoLa~\cite{dola} follows a similar principle, but without external knowledge from other LLMs. DoLa leverages that the knowledge bias is mainly caused by early layers, and utilizes this phenomenon to mitigate hallucinations by contrasting the predictive distributions induced by different layers within one LLM. Naturally, a similar principle can also be applied to VLMs~\cite{vcd, zhang2024debiasing, wang2024ICD, halc}. VCD~\cite{vcd} observes that perturbed images (\eg, by adding Gaussian noise) have an increased tendency to hallucinate (\ie,  the winning logits generated from the perturbed image are more often induced by a language prior).
 Therefore, the final logits are a linear combination of the ones induced by the original image and perturbed image, respectively. VDD~\cite{zhang2024debiasing} follows the same logic as VCD but with an additional calibration step. To be specific, a weight matrix $W$ is learned to transform the predictive distribution produced from the case of replacing the noisy image with a dummy test with no images to be a uniform distribution for each answer. Afterwards, the same criterion as VCD is applied, \ie, the final logit is a linear combination of the calibrated logits with and without the original image. Instruction contrastive decoding (ICD)~\cite{wang2024ICD} extends the contrastive principle to the introductions/prompts literally by adding a prefix (\eg, \texttt{You are a confused object detector}) to the standard prompt, obtaining a contrasted distribution with more hallucinations. The calculation of the final logit is the same as \VCD~and VDD~\cite{zhang2024debiasing}. \PAI~obtains the contrasted distribution via only utilizing textual inputs with additional attention modifications.~\cite{yang2025mitigating} introduces adaptive deactivation, setting the weights of text tokens to zero for certain heads identified as hallucination-prone, based on the ratio of text attention to image attention. Most contrastive decoding methods for hallucination mitigation operate within internal states and require a contrasted distribution from either a distorted visual input~\cite{vcd, zhang2024debiasing}, or a pre-defined layer bucket~\cite{dola}, or prompt engineering~\cite{wang2024ICD}.
 
\begin{table*}[ht]
\centering
\begin{adjustbox}{width=\linewidth,center}
\begin{tabular}{lcccccc}
\toprule
\multirow{2}{*}{\textbf{Methods}}   &  
\multicolumn{6}{c}{\textbf{Free of}} \\
&  pre-defined layers & visual editing  & prompt tuning &  specific decoding & external knowledge & contrastive decoding\\
\midrule
\ICD~  &  \yes & \yes & \no  & \yes  &\yes  &\no \\
\CGD~ & \yes & \yes & \yes & \no & \no     & \yes \\
\VCD~  &  \yes  & \no & \yes & \yes &\yes & \no \\ 
\OPERA~ & \yes & \yes & \yes & \no & \yes & \yes \\
\HALC~  & \no  & \yes & \yes & \yes  &\no  &\no \\
PAI~\cite{pai} & \yes & \yes  & \yes & \yes & \yes & \no \\ 
\textbf{Energy-guided (Ours)}     & \yes & \yes & \yes & \yes & \yes &\yes \\
\bottomrule
\end{tabular}
\end{adjustbox}
 \caption{\emph{Taxonomy Comparison of Object Hallucination Mitigation Methods.} \CGD~and \OPERA~necessitate beam search~\cite{beamsearch} as the decoding method. \CGD~ and \HALC~require to access CLIP~\cite{clip} and GroundingDINO~\cite{groundingdino}, respectively. \VCD~and \PAI~need twice feed-forward passes to obtain the contrastive destitution. Our method requires minimal effort to mitigate object hallucination. }
\label{tab:hallucination-specification}
\end{table*}
  
\paragraph{Non-contrastive decoding in VLMs} Another line of hallucination mitigation methods does not rely on contrasting another logit distribution~\cite{deng2024seeing, huang2023opera, jiang2024devils}. \CGD~ aims to mitigate object hallucination on a sentence level. Particularly, it leverages the powerful vision-language alignment capabilities of CLIP to identify sentences that are better aligned with the corresponding visual embeddings. This ensures that the generated responses not only have higher sentence likelihood but also higher CLIP scores. 
However, its performance gain highly relies on the capability of external models. Further, the possible decoding methods are redistricted to nucleus sampling~\cite{nucleus_sampling} and beam search in order to create the candidate sentences. OPERA~\cite{huang2023opera}
observes that
VLMs tend to generate new tokens by focusing on a few summary tokens but not necessarily taking all the previous tokens into account. Therefore, the hallucination is mitigated by penalizing the ``over-trust'' logit. However, the hysteresis of beam-search necessitates a mechanism named retrospection-allocation, \ie, the decoding procedure may roll back to the identified summary token and select other candidates for the next token prediction except for the candidates selected before. Consequently, \OPERA~ iteratively operates with the beam-search decoding, which results in high-computational demand at the inference stage but also severely restricts its applicable scenarios.~\cite{jiang2024devils}~enhances the visual attention weights of specific layers, selected based on their visual attention ratios. Our method is highly efficient, which only requires one single forward pass to calculate the energy score at each layer.

\paragraph{Latent representations in language models}

Analyzing and understanding the decoding mechanism of transformer-based language decoders has been studied from various perspectives including but not limited to attention maps/patterns~\cite{attention_is_not_only_weight, conmy2023automated,Chefer_2021_ICCV, meng2022locating} and the intermediate representation~\cite{ffn-kv-transformer, transformer_promoting_concepts, analyzing_embedding, logit_lens, halawi2023overthinking,tuned_lens} with the application of early exiting~\cite{logit_lens,halawi2023overthinking} or model knowledge editing~\cite{concept_removal, concepterasure}. Model knowledge editing refers to identifying and removing a (linear) concept subspace from the representation, preventing any (linear) predictor from recovering the concept. Meanwhile, early exiting in the context LLMs refers to projecting the hidden states extracted at each layer to the learned ``unembedding'' matrix of the language decoder. By doing this, one can obtain the multiple logit distributions for the following decoding. 

Unlike existing hallucination mitigation methods such as~\VCD~(which necessitates generating a sophisticated noisy version of the original visual inputs), \OPERA~(which relies on the beam-searching decoding mechanism), \HALC~(requiring a pre-defined layer bucket and an object detector), and \MMVP~(relies on additional fine-tuning), our method is derived through the lens of internal states of a language decoder. Termed energy-guided decoding, it avoids the need of visual distortion, or prompt engineering, or external detectors making it free from contrastive decoding. More importantly, the energy score at each layer can be computed with a single forward pass, making our method significantly less computationally demanding compared to \OPERA~and \HALC~.

 \section{Methods}

\subsection{Vision-Language Model Summary}
Generally, the input tokens processed by VLMs consist of visual and text tokens.
The visual tokens of the input image are denoted by $\{I_1, I_2, I_3, \cdots, I_N \}$, and the corresponding language tokens are denoted by $\{W_1, W_2, W_3, \cdots, W_M \}$. $N$ and $M$ are the corresponding cardinalities of the visual and language tokens, respectively. The visual and language tokens are subsequently concatenated, resulting in a set $\mathbf{x}$ of input tokens with size $T=M+N$. VLMs are commonly trained in an autoregressive manner with a causal attention mask meaning that the prediction of the current token $x_t$ only depends on the previous tokens, formally, 
\begin{align}
\mathbf{h} &= \text{VLM} (\mathbf{x}) = \{h_0, h_1, \cdots, h_{T-1}\},
\end{align}
where $\mathbf{h}$ is the output state of the final layer of LLM decoder, and the size of $h_t$ is $f_\text{dim}$. A learned vocabulary head $\mathcal{H}$ with the size of $V_\text{size}$ is utilized to obtain the logits. The learned vocabulary head $\mathcal{H}$ plays a similar role as the penultimate layer of standard discriminative classifier, formally, 
\begin{align}
    p(x_t|x_{<t}) = \text{Softmax} [\mathcal{H}(h_t)],
\end{align}
where $x_{<t}$ denotes the sequence of tokens before $t$-th position $ \{x_i\}_{i=0}^{t-1}$ and $\mathcal{H} \in \mathbb{R}^{f_{\text{dim}} \times V_\text{size}}$.

\subsection{Empirical Yes Ratio Transfer} 
The source of hallucination appeared in VLMs can be attributed to (i) the embedded knowledge in the language decoder's parameters (\ie, the cause of hallucination in LLMs~\cite{dola}); (ii) a limited capacity of the visual encoder~\cite{mmvp}; (iii) the quality of vision-language instruction-following data~\cite{lure}; (iv) the training objectives~\cite{ouali2024clip}; and (v) the connector that accounts for the feature alignment~\cite{llava_improved}. In this work, we focus on the language decoder in the context of VLMs from the perspective of the yes ratio.

\begin{figure}[tb]
 
    \centering
    \includegraphics[width=0.33\linewidth]{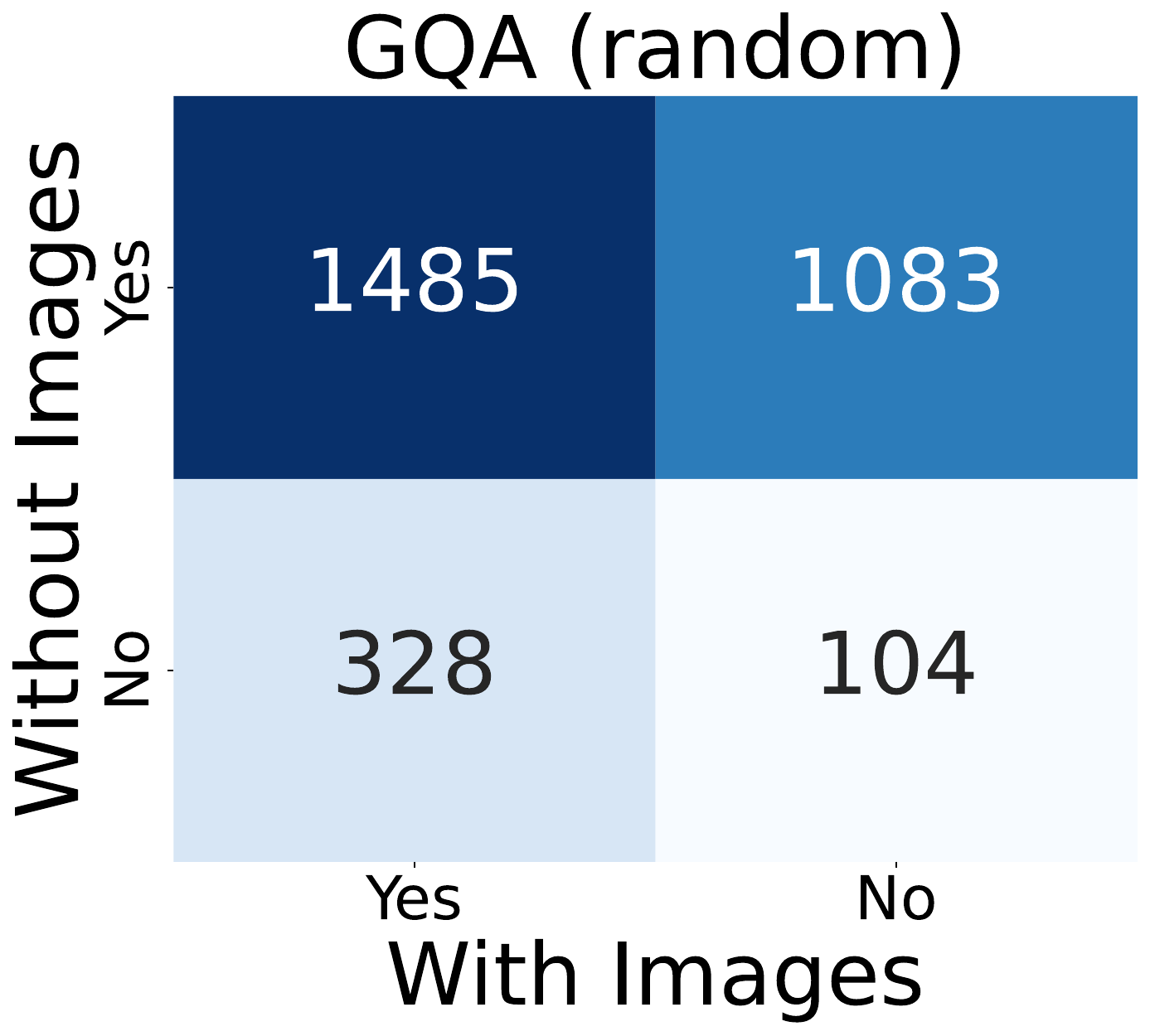}%
     \includegraphics[width=0.33\linewidth]{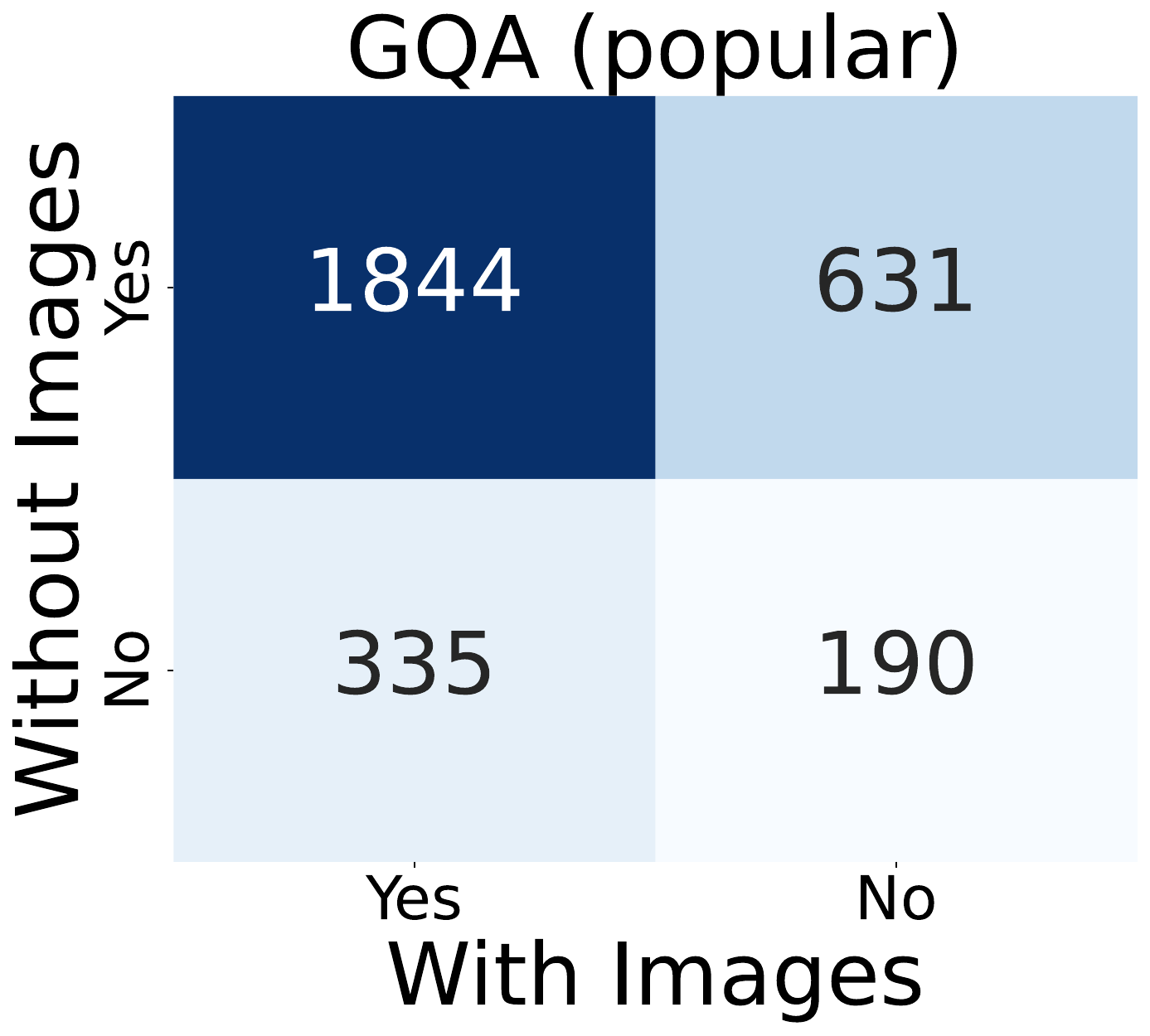}%
      \includegraphics[width=0.33\linewidth]{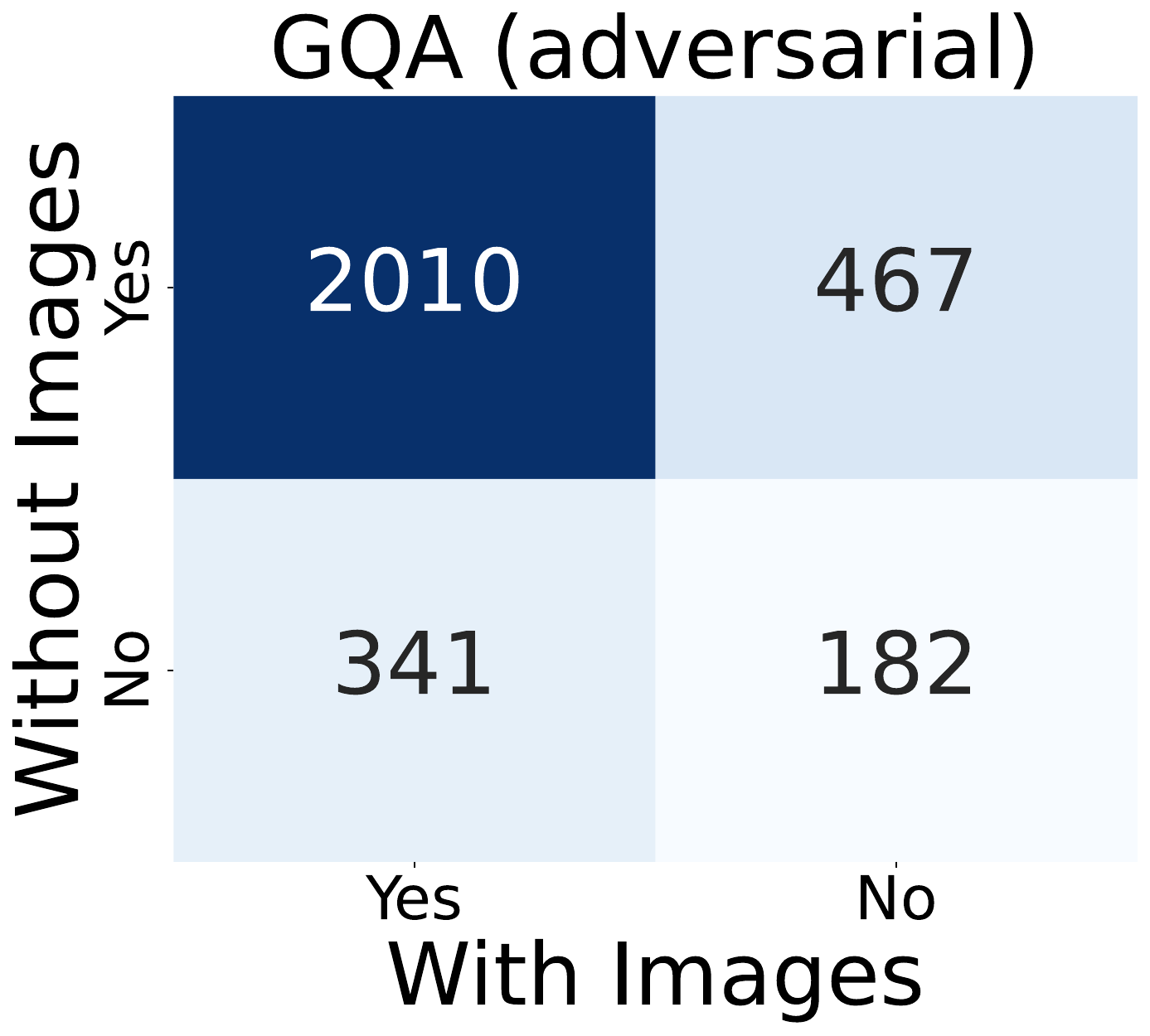}
    \caption{\emph{Transfer of ``Yes'' Ratio} from non-visual input to visual inputs using greedy decoding. Three settings of POPE-GQA~\cite{pope} are utilized including random (left column), popular (middle  column), and adversarial (right column).}
    \label{yes_ratio_greedy}
\end{figure}

\paragraph{Yes ratio transfer} We start with the evaluation covering two scenarios—one that includes visual input and one without visual input.\footnote{The codebase is \url{https://github.com/haotian-liu/LLaVA/tree/main/llava/eval}}
Figure~\ref{yes_ratio_greedy} provides some interesting empirical observations:
\begin{enumerate}
    \item The ``Yes'' ratio---answers labeled ``Yes'' out of total questions---generally increases as tasks grow more challenging, from random to popular to adversarial settings.
    \item The ``Yes'' ratio is initially high without visual input, suggesting that models are biased towards ``Yes'' due to language priors in this dataset.
\end{enumerate}
These findings indicate that the ``Yes'' bias of language models can transfer to VLMs, especially under more challenging, hallucinatory tasks.

\paragraph{Yes/No confidence} We further visualize the confidence of ``Yes'' and ``No'' measuring by the corresponding token probability. One can see from Figure~\ref{yes_ratio_confidence_comparison} (top row) that the model is overconfident to say ``Yes'' (blue area) compared to say ``No'' (orange area) across three settings of POPE-GQA~\cite{gqa}.
Ideally, the confidence levels for both response alternatives are balanced, such as shown in Figure~\ref{yes_ratio_confidence_comparison} (bottom row) (which is actually obtained by our proposed method).

\subsection{Energy-Guided Decoding}
 
The architecture of transformer-based language decoders enables direct decoding from hidden states of each layer into the vocabulary space using the model's pre-trained ``unembedding'' matrix $\mathcal{H}$. This early exiting technique is termed the ``logit lens'' in~\cite{logit_lens} and has for instance been utilized to analyze decoding mechanism of transformer-based language decoders~\cite{analyzing_embedding, transformer_promoting_concepts, ffn-kv-transformer} and to improve factuality of LLMs~\cite{dola}. The distinction between``logit lens'' and energy guided decoding is that the logit lens is a common approach to probe hidden states layer-by-layer in LLMs for analysis, while energy-guided decoding uses energy scores to determine the optimal layer for decoding.

In this work, we aim to identify the layer where the hidden states is the most reliable representation of the input. In particular, we utilize the energy score~\cite{energybased_ood}, which is a frequently used out-of-distribution (OOD) score for cross-entropy trained classifiers to detect OOD samples that are semantically different from the training data. For a pretrained discriminative classifier parameterized by $\theta$ and output label space $\mathcal{Y} \in \{1, 2,  \dots, C\}$, the energy score for an input $\mathbf{x}$ is given by
\begin{align}
   E_\theta(\mathbf{x}) &= -\logsumexp [f_{\theta}(\mathbf{x})]
    := -\log \sum\nolimits_y e^{f_{\theta}(\mathbf{x})[y]},
\nonumber
\end{align}
where $f_\theta(\mathbf{x})$ is the vector of class logits. 
Inputs with higher energy scores are regarded as likely outliers. If a discriminative network is trained appropriately~\cite{JEM}, then the energy score corresponds directly to the negative log-evidence $\log p(\mathbf{x})$. Alternatively, in certain scenarios, the energy score obtained from purely discriminatively trained networks may correspond to some form of feature log-likelihood~\cite{burapacheep2024your}.

\begin{figure}[tb]
    \centering
    \includegraphics[width=0.33\linewidth]{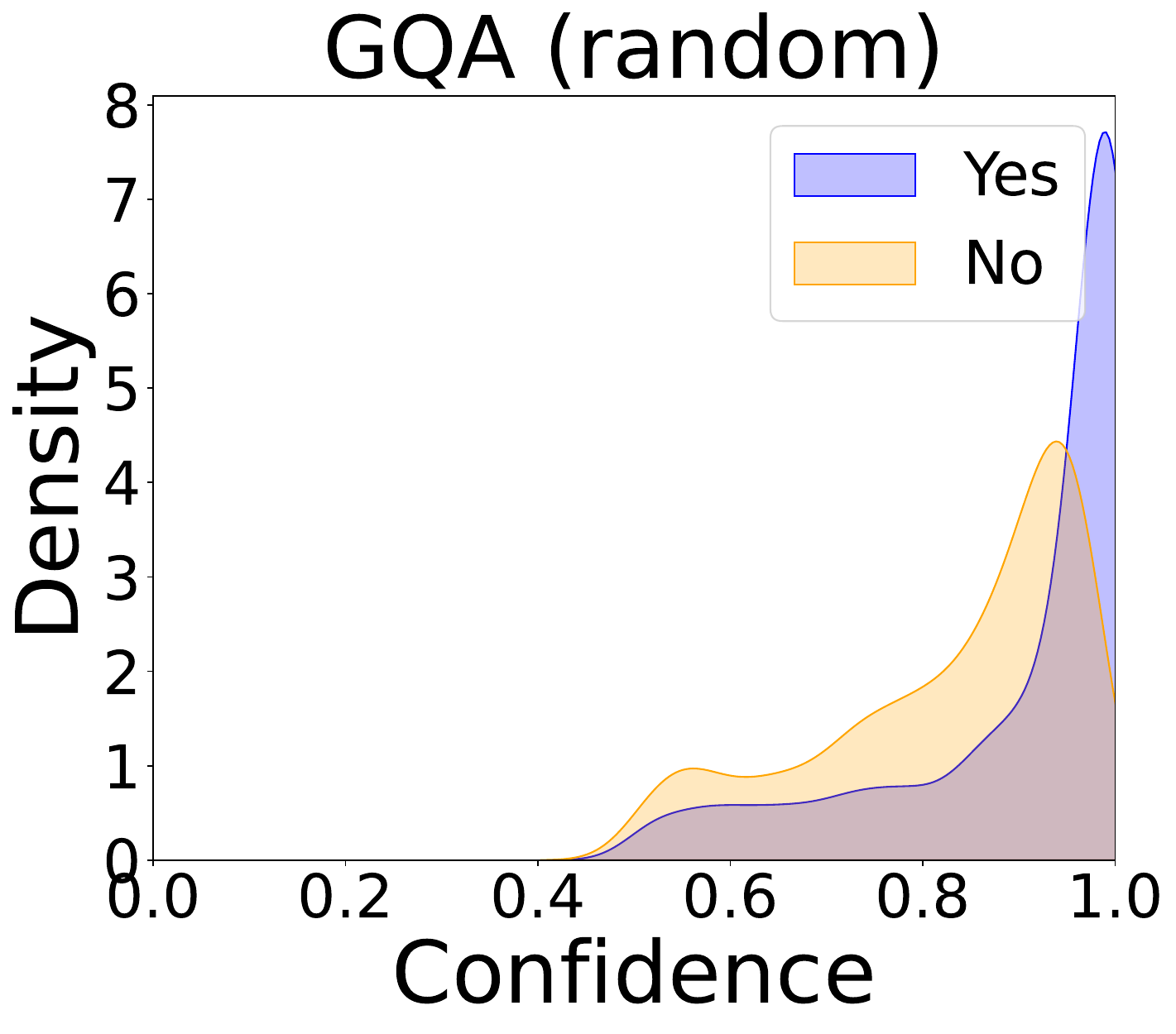}%
     \includegraphics[width=0.33\linewidth]{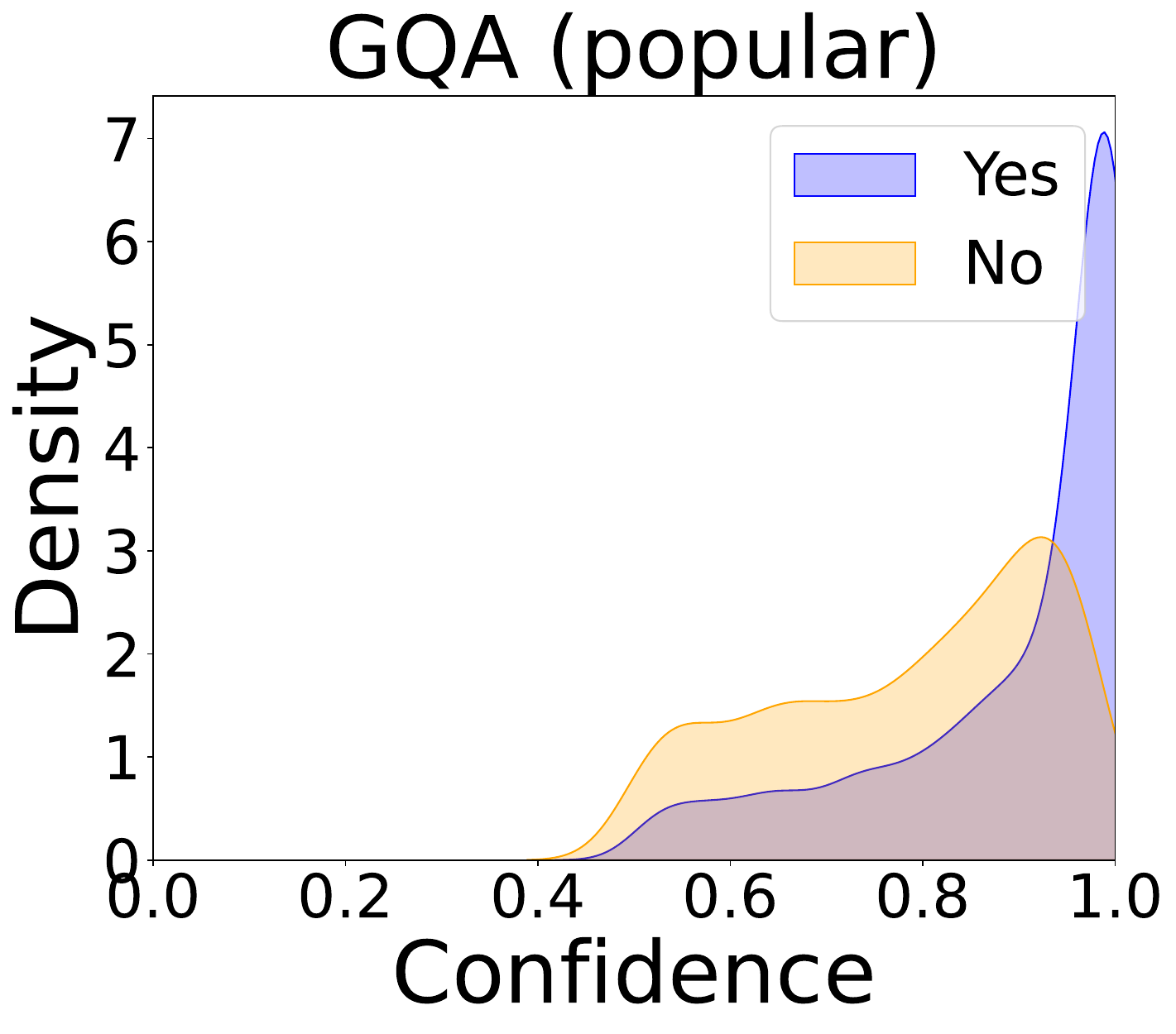}%
      \includegraphics[width=0.33\linewidth]{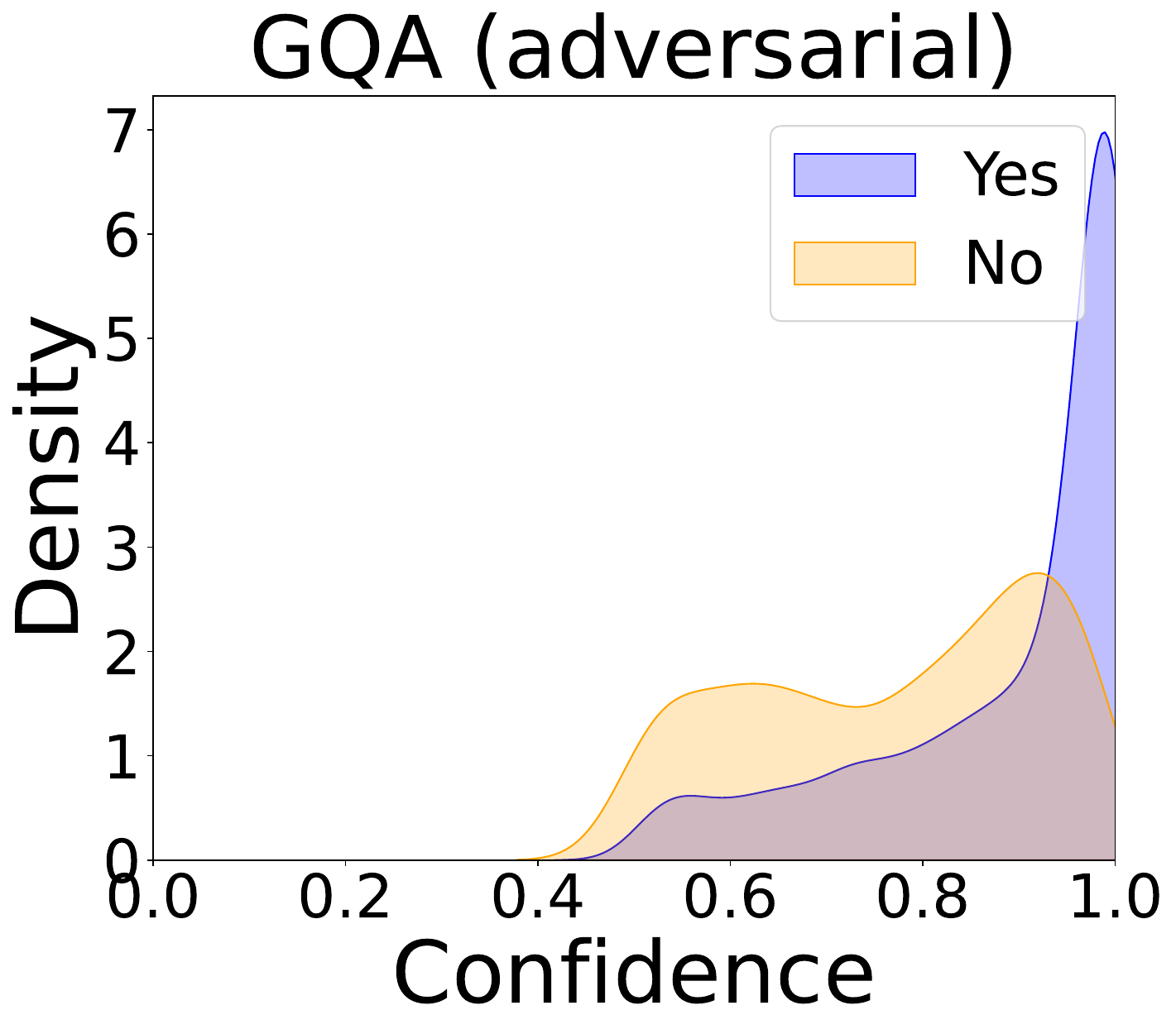}
 
    \centering
    \includegraphics[width=0.33\linewidth]{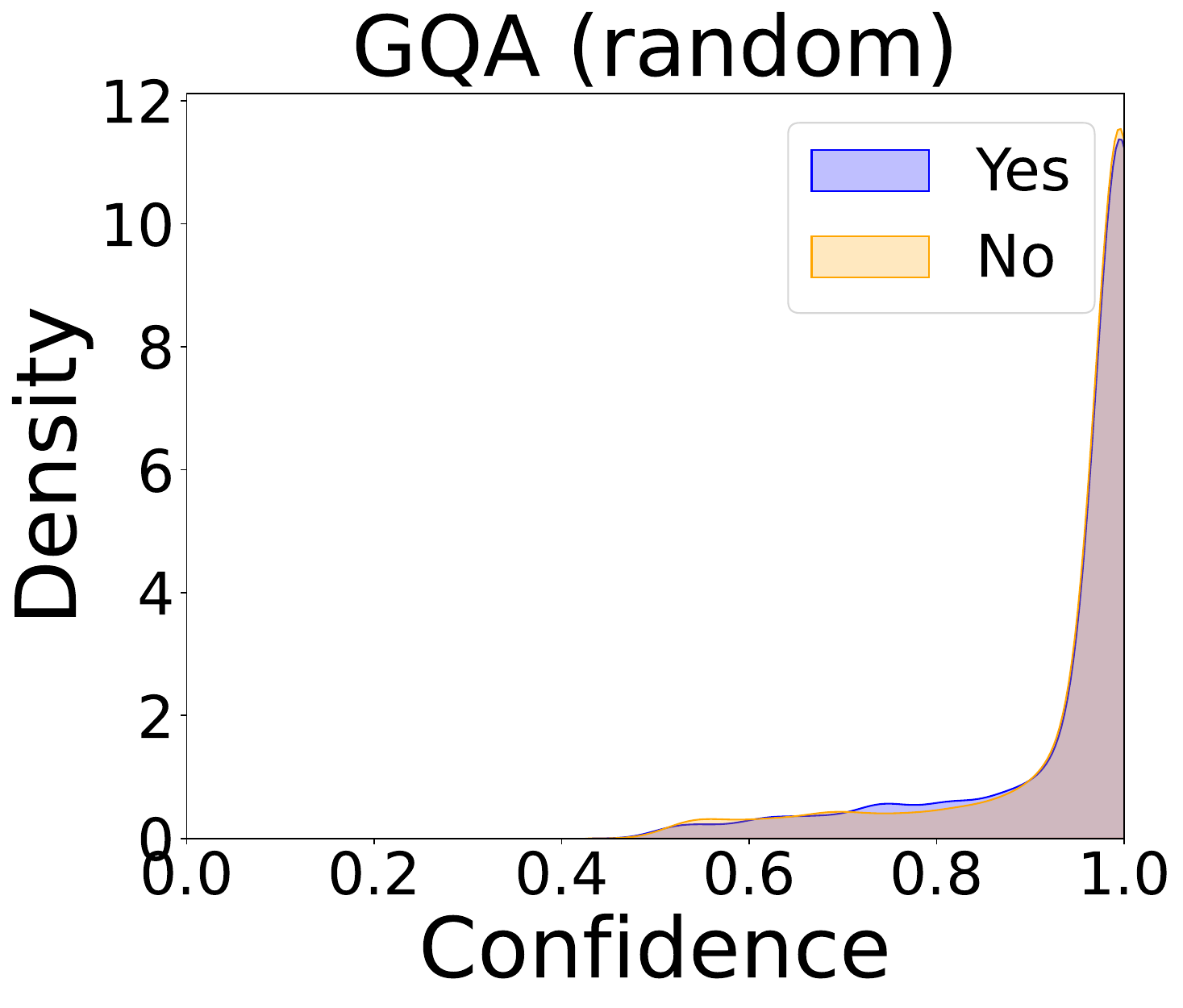}%
     \includegraphics[width=0.33\linewidth]{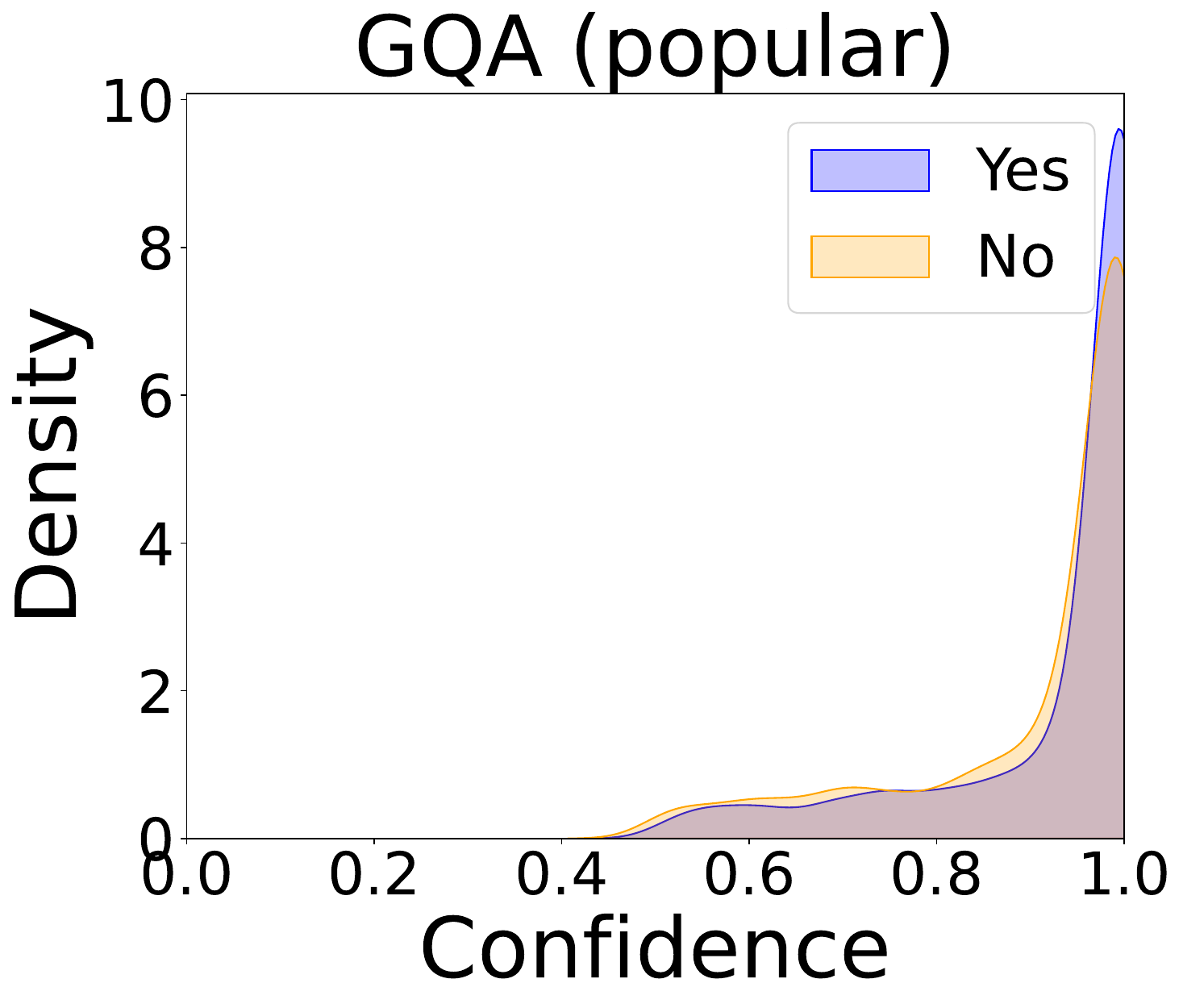}%
      \includegraphics[width=0.33\linewidth]{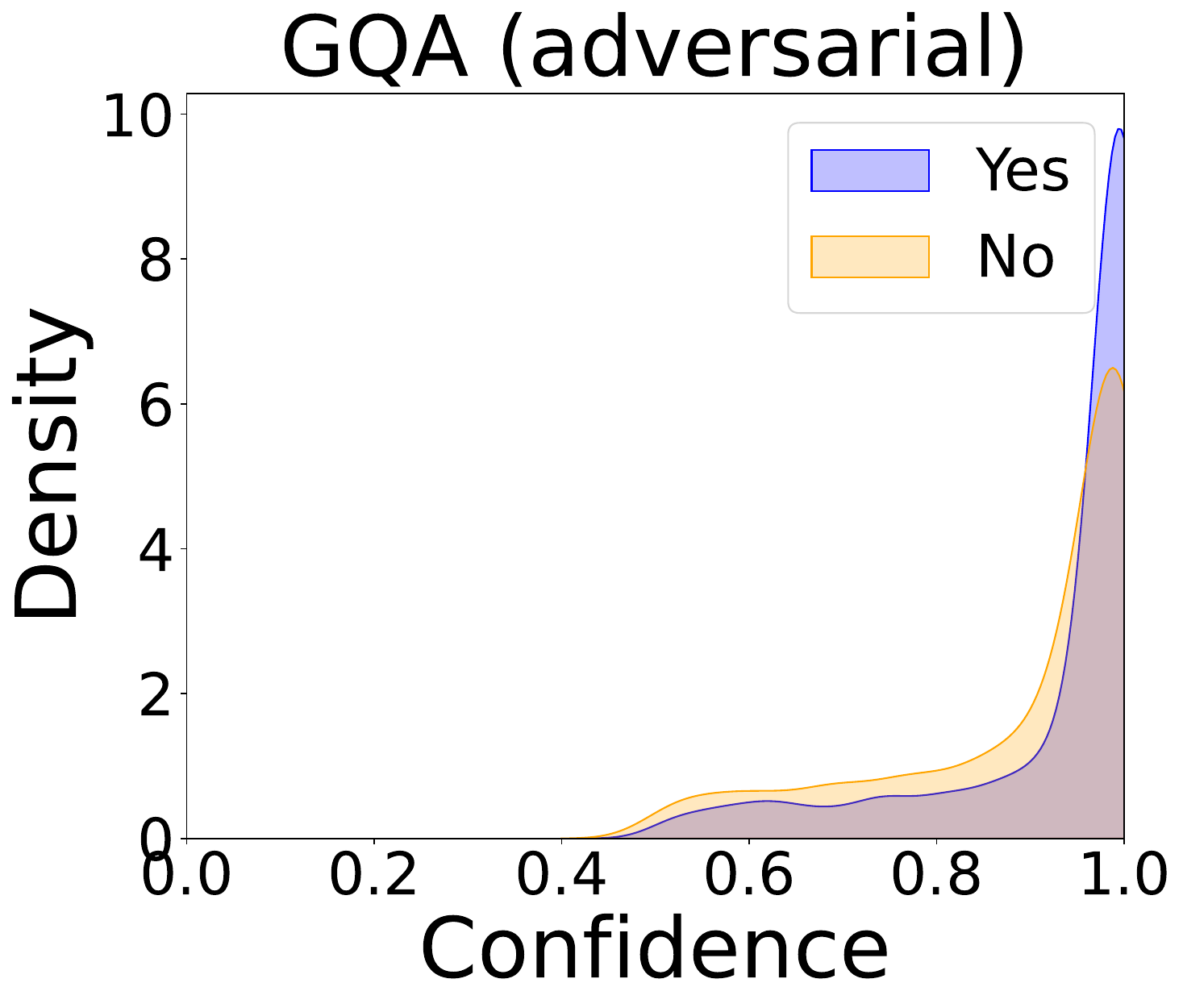}
    \caption{\emph{Kernel Density Estimation} for answers with yes \textcolor{blue}{(blue)} and the ones with no \textcolor{orange}{(orange)}, using hidden states from the last layer for decoding (top row) and the ones with minimum energy for decoding (bottom row), respectively. The confidence is measured by the predictive probability of the corresponding token, \ie, the probability distribution after the  $\softmax$ layer.}
    \label{yes_ratio_confidence_comparison}
\end{figure}

In the context of VLMs, that employ a transformer-based language decoder,
we can interpret the hidden states from each layer in the decoder as different representations of the input. Therefore, we propose to identify the layer where the hidden states (after applying the ``unembedding'' via $\mathcal{H}$) provide the most reliable representation of both visual and textual inputs. Specifically,  we first obtain the corresponding logit vector at each layer by applying the vocabulary head $\mathcal{H}$ to the feature representation.  We then calculate the energy score to identify the layer whose hidden state provides the most reliable representation of the input. Formally, the energy score is given by
\begin{align}
    \textbf{Energy}(h_t^k) = -\logsumexp [\mathcal{H}(h_t^k)]      
\end{align}
where $\mathcal{H}(h_t^k)$ denotes the logit vector (of size $V_{\text{size}}$) calculated at layer $k$ for predicting token $t$. To this end, the layer $k^*=\arg\min_k \textbf{Energy}(h_t^k)$ with the lowest score is consequently used for decoding.
An illustration of our method is shown in Fig.~\ref{fig:toy_example}, and Alg.~\ref{alg: energy_guided_decoding} lists the respective Pytorch-like pseudocode.
Empirically, the use of energy score in this work is motivated by the observations illustrated in Figure~\ref{yes_ratio_confidence_comparison}: negative (``No'') responses are consistently under-confident and a source of the prevalence of ``Yes'' answers in Figure~\ref{yes_ratio_greedy}.
In order to balance this asymmetry in confidence we therefore neutralize differences in the logit vectors by identifying the layer with the most reliable representation (in terms of the energy score) for the subsequent decoding step.

\section{Experiments}
It is critical to maintain the capability of open-ended generation while mitigate hallucination. Therefore, the proposed method is demonstrated on four benchmarks including POPE~\cite{pope}, MME~\cite{mme}, and MMVP~\cite{mmvp}, and CHAIR~\cite{objectHallucination}. We consider three representative open-sourced VLMs with different architecture designs.

\subsection{Models and baselines}
\paragraph{VLM backbones} Three representative VLMs including LLaVA-1.5~\cite{llava_improved}, InstructBLIP~\cite{instructblip}, and \mplugowl~ are employed to evaluate the performance of hallucination mitigation. Specifically, LLaVA-1.5~\cite{llava_improved} simply utilizes a Multilayer perceptron (MLP) layer to align the visual feature and text feature. InstructBLIP~\cite{instructblip} employs the Q-former~\cite{blip2} to extract instruction-aware visual features from the output embeddings of the frozen image encoder. They both employ Vicuna-7B~\cite{vicuna2023} as the language decoder. Different from~\llavaimproved~ and \instructblip~ that utilize a standard language decoder, \mplugowl~ employs a modality-adaptive language decoder to tackle different modalities, and LLaMA-2-7B~\cite{llama2} as the language decoder.
 
\begin{algorithm}[t]
\caption{Energy-Guided Decoding} 
\SetKwInOut{Input}{Input}\SetKwInOut{Output}{Output}
\Input{\PyCode{outputs.hidden\_states} \PyComment{List of hidden states from transformer model}}
\Output{\PyCode{next\_token\_scores} \PyComment{Next-token scores derived from the minimal-energy hidden state}}
    
\PyCode{all\_hidden\_energy = torch.zeros(1, len(outputs.hidden\_states)).to("cuda:0")} \PyComment{Initialize tensor to store energy values}\;

\For{$i \gets 1$ \KwTo \PyCode{len(outputs.hidden\_states)}}{
    \PyComment{Compute logits for current hidden state}\;
    \PyCode{hidden\_logits = self.lm\_head(outputs.hidden\_states[i])} 
    
     \PyComment{Calculate "negated energy score" over the logits of the last token}\;
    \PyCode{hidden\_energy = LogSumExp(hidden\_logits[:, -1, :])} 
    
     \PyComment{Store energy in all\_hidden\_energy}\;
    \PyCode{all\_hidden\_energy[0, i] = hidden\_energy}
    
}
\PyComment{Find the index of the minimal energy}\;
\PyCode{max\_idx = torch.sort(all\_hidden\_energy, descending=True).indices[:, 0]} 

\PyComment{Compute next-token scores from highest energy state}\;
\PyCode{next\_token\_scores = self.lm\_head(outputs.hidden\_states[max\_idx])}  
\label{alg: energy_guided_decoding}
\end{algorithm}

\paragraph{Decoding baselines} To ensure reproducibility\footnote{The codebase is \url{https://github.com/BillChan226/HALC}}, we use greedy decoding as the baseline method. We also include two training-free methods, \ie, VCD~\cite{vcd} and HALC~\cite{halc}. The beam size is 1 for all baseline methods. For the MME and MMVP benchmarks, we also include nucleus sampling~\cite{nucleus_sampling} as an additional baseline. We use their suggested hyperparameters for VCD~\cite{vcd} and HALC~\cite{halc}. A detailed hyperparameter settings can be found in supplementary material. More methods such as PAI~\cite{pai} and \OPERA~ can also be found in supplementary material.  
\subsection{Datasets and Evaluation Metrics}
 
\paragraph{POPE}
Polling-based Object Probing Evaluation (POPE)~\citep{pope} is a common benchmark to evaluate the performance of object hallucination~\citep{vcd, huang2023opera,halc}. It consists of  three datasets including MSCOCO~\cite{mscoco}, A-OKVQA~\cite{a-okvqa}, and GQA~\cite{gqa}. For each dataset, 500 images are sampled with three different sampling strategies, including random sampling, popular sampling, and adversarial sampling. Random sampling refers to randomly drawing objects that do not exist in the image. Popular sampling selects the top half of the most frequent objects in the whole datasets. Adversarial sampling is the most difficult one, where first all objects are sorted based on their co-occurring frequencies with the ground-truth objects, followed by sampling from the most frequent ones not appearing in the image. There are 6 questions for each image, which yields 27,000 query-answer pairs in total.

\paragraph{MME} The original Multimodal Large Language Model Evaluation (MME) benchmark consists of 10 perception-related tasks and 4 cognition-based tasks. We closely follow~\cite{yin2023woodpecker,vcd, halc} to perform hallucination evaluation on the perceptional subtasks. Specifically, the existence and count tasks are employed for object-level hallucination evaluation and the position and color tasks for attribute-level hallucination evaluation. Each image is designed with two questions.

\paragraph{MMVP} Multimodal Visual Patterns (MMVP)
benchmark~\cite{mmvp} consists 150 images with 300 questions. The collected paired images are CLIP-blind meaning that their cosine similarity exceeds 0.95 for CLIP embeddings and less than 0.6 for DINOv2 embeddings.

\begin{table}[t]
\centering
\begin{adjustbox}{width=0.95\linewidth,center}
\begin{tabular}{llc}
 \toprule
   Datasets  & Hallucination types & \# Pairs \\ 
\midrule

POPE-MSCOCO~\cite{mscoco}  & category         & 9,000\\
POPE-AOKVQA~\cite{a-okvqa} & category     & 9,000\\
POPE-GQA~\cite{gqa} & category& 9,000\\  
 
MME~\cite{mme} & category, attribute & 240\\  
MMVP~\cite{mmvp} &category, attribute, relation & 300 \\
\bottomrule
\end{tabular}
\end{adjustbox}
\caption{Specifications of hallucination benchmarks.}
\label{tab:hallucination}
\end{table}

\paragraph{Evaluation Metrics}
We closely follow the evaluation protocol established by VCD~\cite{vcd} for the POPE benchmark and MME benchmark. Specifically, accuracy and F1 score (\ie, the harmonic mean of precision and recall) are commonly employed to measure the presence of hallucinations. Unlike \VCD{}, which simply shows the values of the yes ratio, we choose to depict the gap between the the predicted and the expected yes ratio, which reflects the degree of bias more directly. To be specific, the yes-ratio gap is defined as  
\begin{align}
    \Delta_\text{gap} = \left| \frac{\# \text{ of answers with yes}} {\# \text{ of total questions}}-0.5 \right|,
    \label{eq:gap_definition}
\end{align} 
where $|\cdot|$ denotes the absolute value and 0.5 represents the expected yes ratio because the dataset is balanced.
For the MME benchmark, we follow \VCD~ to report the sum of accuracy (\ie, the number of correct answers over the total questions) and accuracy$+$ (the number of correctly answering both questions given one image over the total number of images) as the final score. For the MMVP benchmark, we report the same metrics as \POPE~.

\subsection{Experimental results}

\begin{table*}[ht]
\begin{adjustbox}{width=0.9\linewidth,center}
\begin{tabular}{clllllllllll}
\toprule
\multirow{2}{*}{\textbf{Datasets}}       & \multirow{2}{*}{\emph{Settings}}   & \multirow{3}{*}{\textbf{Methods}} & \multicolumn{3}{l}{\textbf{LLaVA-1.5}~\cite{llava_improved}}
& \multicolumn{3}{l}{\textbf{InstructBLIP}~\cite{instructblip}} & \multicolumn{3}{l}{\textbf{mPLUG-Owl2}~\cite{mplugowl2}}  \\

& & & Acc.$\uparrow$ &  F1 Score$\uparrow$ & $\Delta_\text{gap}\downarrow$  & Acc.$\uparrow$  & F1 Score$\uparrow$ & $\Delta_\text{gap} \downarrow$  & Acc.$\uparrow$  & F1 Score$\uparrow$ & $\Delta_\text{gap} \downarrow$      \\
\midrule
\midrule
\multirow{15}{*}{\rotatebox{90}{MSCOCO}}      & \multirow{5}{*}{\textit{Random}}  
& Greedy & \best{89.37} &  \best{89.33} & \best{0.37}   &\best{90.17}&\best{89.86}&3.03 &83.30&84.76&9.57                   \\&                                                             
& VCD  &84.83 &85.30& 3.17   &84.47&84.38& \best{0.53} &80.83&82.67&10.57                    \\& 
& \HALC~ &\second{89.30}& \second{89.25}& \second{0.50} &\second{89.73}&\second{89.52}&\second{2.07} &\second{83.57}& \second{84.96}&\second{9.30}                      \\&            
                                         
&\cellcolor{mygray}Energy (Ours) &\cellcolor{mygray}87.50 &\cellcolor{mygray}86.22&\cellcolor{mygray}9.30  &\cellcolor{mygray}86.80&\cellcolor{mygray}85.10&\cellcolor{mygray}11.40&\cellcolor{mygray}\best{87.73}&\cellcolor{mygray}\best{86.55}&\cellcolor{mygray}\best{8.80} \\ \cline{2-12} \noalign{\smallskip} 

& \multirow{4}{*}{\textit{Popular}}                           
& Greedy &\second{86.00} & \second{86.41}&3.00  &\second{83.47}& \best{84.05}& \best{3.67} &77.40&80.43&15.47       \\  &                                                             
& \VCD~ &81.77 &82.81& 6.10 &77.73&79.12& 6.67&75.07&78.57&16.33                   \\        & 
& \HALC~  &\best{86.10} &\best{86.47}& \best{2.70} & 82.30&83.20&\second{5.37}  &\second{77.57}&\second{80.54}&\second{15.30}    \\ &                                                             
&\cellcolor{mygray}Energy (Ours) &\cellcolor{mygray}85.80&\cellcolor{mygray}84.63&\cellcolor{mygray}7.60&\cellcolor{mygray}\best{83.70}&\cellcolor{mygray}82.22&\cellcolor{mygray}8.31 &\cellcolor{mygray}\best{86.50}&\cellcolor{mygray}\best{85.39}&\cellcolor{mygray}\best{7.57} \\ \cline{2-12} \noalign{\smallskip}  
                         
& \multirow{4}{*}{\textit{Adversarial}}                       
& Greedy  &79.10 &80.96& 9.77  &\second{80.67}& \best{81.82}& \best{6.33}  &73.80&77.98&19.00   \\&
& VCD  &76.17 &78.73&12.03 & 75.87&77.94& 9.40   &72.80&77.01&\second{18.33}                  \\&                                              
& HALC  &\second{79.27} &\second{81.05}&  \second{9.40}  &79.47&\second{80.99}&8.00 &\second{74.00}&\second{78.11}&18.80                     \\&    
& \cellcolor{mygray}Energy (Ours) &\cellcolor{mygray}\best{82.90} &\cellcolor{mygray}\best{82.03}&\cellcolor{mygray}\best{4.83}&\cellcolor{mygray}\best{82.17}&\cellcolor{mygray}80.90& \cellcolor{mygray}\second{6.64} &\cellcolor{mygray}\best{84.50}&\cellcolor{mygray}\best{83.56}&\cellcolor{mygray}\best{5.70}\\  \midrule

\multirow{12}{*}{\rotatebox{90}{A-OKVQA}}     

& \multirow{4}{*}{\textit{Random}}      
& Greedy &85.70 &86.90&9.17 &\best{89.13}&\best{89.50}& \best{3.47}  &79.23&82.45&18.30               \\&                                                           
& VCD &80.77 &82.85&12.17 &83.23&84.23&6.30  &77.50&80.94&\second{18.03}               \\    &                                                 
& HALC&\second{85.80} &\second{86.98}&\second{9.07}  &88.27&88.85&\second{5.20}  &\second{79.40}&\second{82.55}&18.07                    \\       &                                                             
& \cellcolor{mygray}Energy (Ours) &\cellcolor{mygray}\best{88.60} &\cellcolor{mygray}\best{88.30}&\cellcolor{mygray}\best{2.6}&\cellcolor{mygray}\second{89.07}&\cellcolor{mygray}\second{88.44}&\cellcolor{mygray}5.40 &\cellcolor{mygray}\best{87.97}&\cellcolor{mygray}\best{87.36}&\cellcolor{mygray}\best{4.77}\\ \cline{2-12} \noalign{\smallskip} 

& \multirow{4}{*}{\textit{Popular}}                           
& Greedy &79.90 &82.52&14.97 &\second{79.57}&\second{81.92}&\second{13.03}  &71.83&77.59&25.70    \\&                                 
& VCD &76.47 &79.83&16.67  &76.87&79.73&14.13  &71.40&76.97&\second{24.20}                   \\   & 
& HALC & \second{79.97} & \second{82.56}& \second{14.9} &78.20&81.09&15.27  &\second{72.17}&\second{77.79}&25.30                \\ &                                                           
&\cellcolor{mygray}Energy (Ours) &\cellcolor{mygray}\best{84.67} &\cellcolor{mygray}\best{84.87}&\cellcolor{mygray}\best{1.33} & \cellcolor{mygray}\best{84.03}&\cellcolor{mygray}\best{83.97} &\cellcolor{mygray}\best{0.37} &\cellcolor{mygray}\best{83.67}&\cellcolor{mygray}\best{83.59}&\cellcolor{mygray}\best{0.47}\\  \cline{2-12} \noalign{\smallskip}

& \multirow{4}{*}{\textit{Adversarial}}                       
& Greedy&69.07 &75.41&25.80 &\second{71.43}&\second76.42&21.17 & 64.80&73.48&32.73     \\&                                                             
& VCD  &68.47 &74.54&23.87 &69.23&74.28&\second{19.63} &\second{65.40}&73.38&\second{30.00}                   \\ &   
& HALC  & \second{69.23} &\second{75.51}&\second{25.63}&70.33&75.91&23.13 &64.90&\second{73.52}&32.57                      \\&                                                             
& \cellcolor{mygray}Energy (Ours) &\cellcolor{mygray}\best{77.40} &\cellcolor{mygray}\best{79.19}&\cellcolor{mygray}\best{8.59 }&\cellcolor{mygray}\best{76.70}&\cellcolor{mygray}\best{78.22}&\cellcolor{mygray}\best{6.96} &\cellcolor{mygray}\best{77.37}&\cellcolor{mygray}\best{78.61}&\cellcolor{mygray}\best{5.83} \\ \midrule

\multirow{12}{*}{\rotatebox{90}{GQA}}      
& \multirow{4}{*}{\textit{Random}}          

& Greedy &85.77 &87.09&10.23  &\best{86.90}&\best{87.30}&\best{3.17}&\second{83.17}&\best{84.88}&11.37                     \\  & 
& VCD &81.33& 83.34&12.07 &80.90&82.07&6.49 &80.77&82.54&\second{10.17}             \\&
& HALC& \second{85.90}& \second{87.19}&\second{10.10} &85.97&\second{86.55}&\second{4.37}   &82.81&\second{84.57}&11.54                      \\ &
&\cellcolor{mygray} Energy (Ours)&\cellcolor{mygray}\best{89.37} &\cellcolor{mygray}\best{89.19}& \cellcolor{mygray}\best{1.63}&\cellcolor{mygray} \second{86.53}&\cellcolor{mygray}85.54&\cellcolor{mygray}6.87 &\cellcolor{mygray}\best{85.60}& \cellcolor{mygray}84.12&\cellcolor{mygray}\best{9.33} \\ \cline{2-12} \noalign{\smallskip} 


& \multirow{4}{*}{\textit{Popular}}                           
& Greedy &74.73 &79.16 & \second{11.27} &\second{76.37}&\second{79.21}&\second{13.7}  &73.77&78.28&20.77    \\&  
& VCD &71.53 &76.82&22.8&73.00&76.32&14.0  &72.03&76.48&\second{18.90}                 \\ &  
& HALC  &\second{74.87}&\second{79.25}&21.13 &74.50&77.99&15.83 &\second{74.13}&\second{78.49}&20.27                 \\ & 
&\cellcolor{mygray}Energy (Ours)&\cellcolor{mygray}\best{82.53} &\cellcolor{mygray}\best{83.40}& \cellcolor{mygray}\best{5.2} & \cellcolor{mygray}\best{80.27}&\cellcolor{mygray}\best{80.15}&\cellcolor{mygray}\best{0.60}&\cellcolor{mygray}\best{79.50}&\cellcolor{mygray}\best{78.82}&\cellcolor{mygray}\best{3.23}\\ \cline{2-12} \noalign{\smallskip}  


& \multirow{4}{*}{\textit{Adversarial}}                       
& Greedy&69.43 &75.85 & 26.57 &\second{71.50}&\second{75.96}&18.56  &69.77&75.77&24.77     \\   &
& VCD  &68.97 &75.14&\second{24.83}&69.10&73.85&\second{18.16} &70.03&75.30&\second{21.30}                \\  & 
& HALC  & \second{69.53} &\second{75.91}&26.47 &69.70&74.88&20.63 &\second{70.13}&\second{75.97}&24.27                \\ &
&\cellcolor{mygray}Energy (Ours)&\cellcolor{mygray}\best{79.63} &\cellcolor{mygray}\best{81.16}& \cellcolor{mygray}\best{8.09} &\cellcolor{mygray}\best{76.57}&\cellcolor{mygray}\best{77.27}& \cellcolor{mygray}\best{3.10}&\cellcolor{mygray}\best{78.27}&\cellcolor{mygray}\best{77.82}&\cellcolor{mygray}\best{2.00}\\

\bottomrule
\end{tabular}
\end{adjustbox} 
\caption{ \emph{Results on POPE Benchmark}. Three representative models include~\llavaimproved~, \instructblip~, and \mplugowl~as the VLM backbones. 
Higher accuracy and F1 score indicate better performance and fewer hallucinations. Lower yes-ratio gap,
$\Delta_\text{gap}$~\eqref{eq:gap_definition},
implies the model is better calibrated. The best performing method within each setting in \best{bold}, the 2nd best is \second{\text{underlined}}. The maximum new token is set to be \textbf{16}. The detailed results regarding precision and recall can be found in the supplementary material. }
\label{tab:pope-vcd-datatlist}
\end{table*}

\paragraph{POPE results} The results in terms of accuracy, F1 score, and yes-ratio gap on POPE benchmark with three datasets including MSCOCO~\cite{mscoco}, A-OKVQA~\cite{a-okvqa}, and GQA~\cite{gqa} are presented in Table~\ref{tab:pope-vcd-datatlist}. First, it is worthwhile to note that our method consistently obtains the highest accuracy and the lowest yes-ratio gap on two datasets including the A-OKVQA~\cite{a-okvqa} and GQA~\cite{gqa} across three different POPE settings with LLaVa-1.5 as the VLM backbone. Specifically, our method outperforms the baseline method greedy with a large margin up to 10.2$\%$ in terms of accuracy and $5.31\%$ in terms of F1 score. More importantly, when the POPE setting is changed from \textit{random} setting to \textit{adversarial} setting meaning when the task difficulty progressively increases, our method maintains the performance gain in terms of both accuracy and F1 score. Further, the effectiveness of our method in terms of accuracy remains when using a less advancing VLM, \ie, InstructBLIP~\cite{instructblip}. Notably, our method consistently achieves the best performance in terms of both accurcy and the yes-ratio gap when using~\mplugowl~. Additionally, we further visualize the confidence of saying ``Yes'' and saying ``No'' before and after using energy-guided decoding in Figure~\ref{yes_ratio_confidence_comparison}. It is evident that greedy decoding (first row) tends to be more confident in saying ``Yes'' than in saying ``No''. More importantly, the confidence in saying ``No'' decreases even further as the setting shifts from random to adversarial. In contrast, our method maintains a similar level of confidence in both saying ``Yes'' and ``No'' even as the task becomes more difficult, transitioning from a random to an adversarial setting.

\paragraph{MME-subset results} The results, including four baselines across two architectures, are presented in Table~\ref{tab:mme-subset}. The reported scores represent the sum of accuracy (\ie, the number of correct answers over the total questions) and accuracy$+$ (the number of instances in which both questions associated with an image are answered correctly, over the total number of images). Our method demonstrates effectiveness in mitigating hallucinations at both the category and attribute levels. Specifically, it effectively reduces hallucinations related to \emph{Count} at the category level and \emph{Color} at the attribute level. This effectiveness suggests that our method (energy-guided decoding) could address the inherent biases built in language decoder. In contrast, all baseline methods obtain relatively lower \emph{Position} score under two different VLMs, indicating that the VLMs are incapable at reasoning tasks. However, LLaVA-1.5 equipped with energy-guided decoding (our method) achieves a noticeable improvement at the \emph{Position} score.

\paragraph{MMVP-subset results} The original MMVP requires either manually checking the generated response or GPT-grader for evaluation. To provide an accurate evaluation, we select 122 image-questions pairs that share the similar prompt template to POPE~\cite{pope} and MME~\cite{mme}, where the initial response from VLMs is a simple ``Yes'' or ``No'', followed by an explanation. Therefore, we can employ the metrics including accuracy and F1 score. The same metrics utilized in \POPE~ including accuracy, F1 score, and yes-ratio gap are reported. One can see from Table~\ref{tab:mmvp-subset} that energy-guided decoding (our method) consistently obtains the best accuracy and yes ratio gap across two architectures. Specifically, the average yes ratio gap is reduced by a margin of 15.59$\%$ compared to the greedy decoding. 
\begin{table}[h]
\centering
\resizebox{\linewidth}{!}{%
\begin{tabular}{llllllc}
\toprule
\multirow{2}{*}{\textbf{Models}}        & \multirow{2}{*}{\textbf{Methods}} & \multicolumn{2}{c}{\textbf{Category-level}}                                   & \multicolumn{2}{c}{\textbf{Attribute-level}}                               & \multicolumn{1}{c}{\multirow{3}{*}{Total Scores$\uparrow$}} \\
                              &                           & \multicolumn{1}{c}{\textit{Existence}$\uparrow$} & \multicolumn{1}{c}{\textit{Count}$\uparrow$} & \multicolumn{1}{c}{\textit{Position}$\uparrow$} & \multicolumn{1}{c}{\textit{Color}$\uparrow$} & \multicolumn{1}{c}{}                       \\ \midrule
\multirow{5}{*}{\llavaimproved~}  & Nucleus   & 180.00 & 86.67 & 75.00 & 135.00 & 476.67\\
                           & Greedy   &  190.00 & 110.00 & 96.67 & 135.00 & 531.67 \\
                           & VCD & 170.00 & 103.33 & 100.00 & 130.90 & 504.23 \\
                            & HALC   &  190.00 & 110.00 & 96.67 & 135.00 & 531.67 \\
                           
                           & \cellcolor{mygray}\textbf{Energy (Ours)} &  \cellcolor{mygray}\best{195.00} & \cellcolor{mygray}\best{148.33} & \cellcolor{mygray}\best{128.33} & \cellcolor{mygray}\best{170.00} & \cellcolor{mygray}\best{641.67}\\
                           \midrule 

\multirow{5}{*}{\instructblip}  & Nucleus & 183.33 & 101.67 & \best{85.00} & 88.33 & 458.33   \\
                               & Greedy & 185.00 & 93.33& 76.67 & 110.00 & 465.00 \\
                               & VCD & 173.33 & 91.67 & 78.33 & 88.33& 431.66 \\
                               &  HALC & \best{185.00} & 81.67 & 70.00 & 110.00 & 446.67 \\ 
                           
                             & \cellcolor{mygray}\textbf{Energy (Ours)} & \cellcolor{mygray}180.00 &\cellcolor{mygray}\best{146.67} & \cellcolor{mygray}56.67 & \cellcolor{mygray}\best{140.00} &\cellcolor{mygray}\best{523.34} \\ 
                           \bottomrule
\end{tabular}
}
\caption{\emph{Results on the MME Dataset.} 
The best performing method within each setting in \best{bold}. The number of maximum new token is set to be \textbf{16}.}
\label{tab:mme-subset}
\end{table}
\begin{table}[tbh]
\begin{adjustbox}{width=\linewidth,center}
\begin{tabular}{lllllll}
\toprule
 Model    & \textbf{Decoding}  & Accuracy$\uparrow$ & Precision & Recall & F1 Score$\uparrow$  & $\Delta_\text{gap} \downarrow$ \\ 
 \toprule
 \multirow{4}{*}{\llavaimproved}  
 & Nucleus &59.02&55.91&85.25&67.53&26.23\\
  & Greedy &57.38&54.46&90.16&\second{67.90}&32.79\\
 & VCD &60.66&56.19&96.72&\best{71.08}&36.07 \\  
  & \cellcolor{mygray}\textbf{Energy (Ours)}  & \cellcolor{mygray}\best{64.75} & \cellcolor{mygray}62.50 &  \cellcolor{mygray}73.77 & \cellcolor{mygray}67.67 & \cellcolor{mygray}\best{9.02} \\
\midrule
 \multirow{4}{*}{\instructblip}   
  & Nucleus & 55.74 & 55.07 & 62.30 & 58.46 & 6.55 \\
  & Greedy & 63.93 & 61.04 & 77.09& \best{68.12} & 13.15  \\
  & VCD &52.46&52.05&62.30&56.72&9.84                                          \\
  &\cellcolor{mygray}\textbf{Energy (Ours)}  & \cellcolor{mygray}\best{64.75} & \cellcolor{mygray}63.24 & \cellcolor{mygray}70.49 & \cellcolor{mygray}\second{ 66.67} & \cellcolor{mygray}\best{5.74}  \\
\bottomrule
\end{tabular}
\end{adjustbox} 
\caption{\emph{Results on MMVP dataset.}  
Higher accuracy and F1 score indicate better performance and fewer hallucinations. Lower yes-ratio gap,
$\Delta_\text{gap}$~\eqref{eq:gap_definition},
implies the model is better calibrated.  
The best entries within each setting are in \textbf{bold}. The number of maximum new token is set to be \textbf{16}.}
\label{tab:mmvp-subset}
\end{table}

 \paragraph{CHAIR Evaluation} It is equally important to maintain the capability of open-ended generation while mitigate hallucination. Therefore, we also evaluate various baseline methods across two backbones with the CHAIR score (\ie, $\text{CHAIR}_I$ and $\text{CHAIR}_S$) as the primary metric. One can see from Table~\ref{tab:CHAIR evaluation} that energy-guided decoding achieves the best performance in terms of CHAIR$_S$. However, our method are less effective in terms of $\text{CHAIR}_I$. One possible reason is that only the hidden states from the last layer directly receive supervision during training.  A detailed description of CHAIR evaluation can be found in supplementary material.

 \begin{table}[h]
 
\begin{adjustbox}{width=0.9\linewidth,center}
 \begin{tabular}{llllll}
  \\[-2em]
 \toprule
  \multirow{2}{*}{Methods}   &  \multicolumn{2}{c}{\textbf{LLaVA-1.5}~\cite{llava_improved}}    & \multicolumn{2}{c}{\textbf{InstructBLIP}\cite{instructblip}} \\
  &  $\text{CHAIR}_S^\downarrow$ & $ \text{CHAIR}_I^\downarrow$  &$\text{CHAIR}_S^\downarrow$ & $ \text{CHAIR}_I^\downarrow$\\ 
\midrule 
 Greedy  & 17.8 & \best{5.4} & \second{26.8} & 13.3  \\
 \VCD~   & 20.6 & 7.1  & 30.6 & \second{11.8}  \\
\HALC~ & \second{17.0} & \second{5.6} & 28.7 & \best{10.5}  \\
\PAI~   &  21.2 & 6.7  & - & -\\                        
\rowcolor{mygray}  
Energy(Ours) & \best{12.2} & 9.4 & \best{21.4} & 14.4  \\
\bottomrule
\end{tabular}
\end{adjustbox}
 
\caption{ \footnotesize{\emph{Open-Ended Generation on MSCOCO Dataset.} The number of maximum new token is set to be \textbf{64}. The utilized prompt is \emph{``Please describe this image in detail''}. The number of sampled images is 500. The best performing method is in \best{bold}. `-' indicates that no code is available.}} 
\label{tab:CHAIR evaluation}
 
\end{table}

\subsection{Ablation studies}

\begin{figure}[h]
    \centering
    \includegraphics[width=0.48\linewidth]{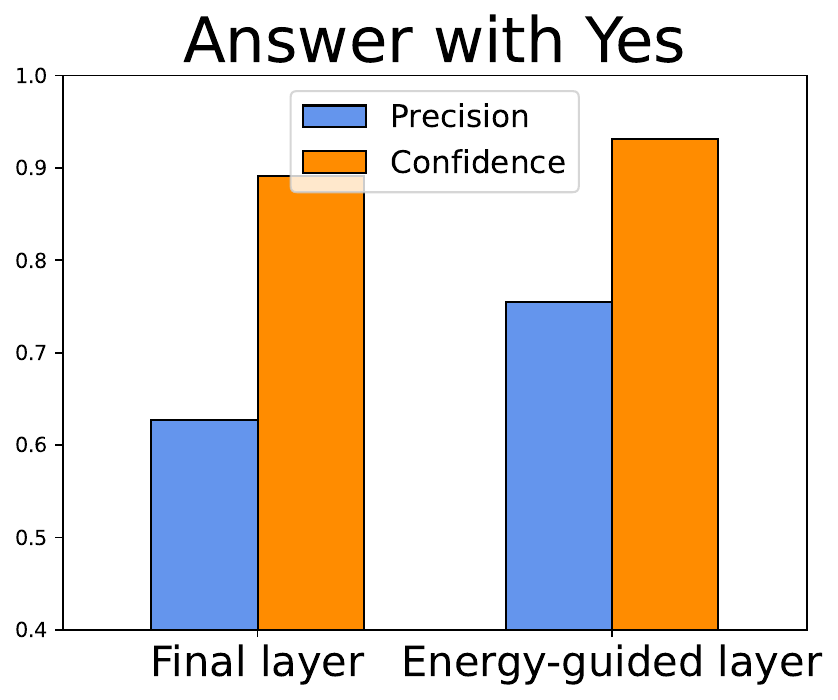}
     \includegraphics[width=0.48\linewidth]{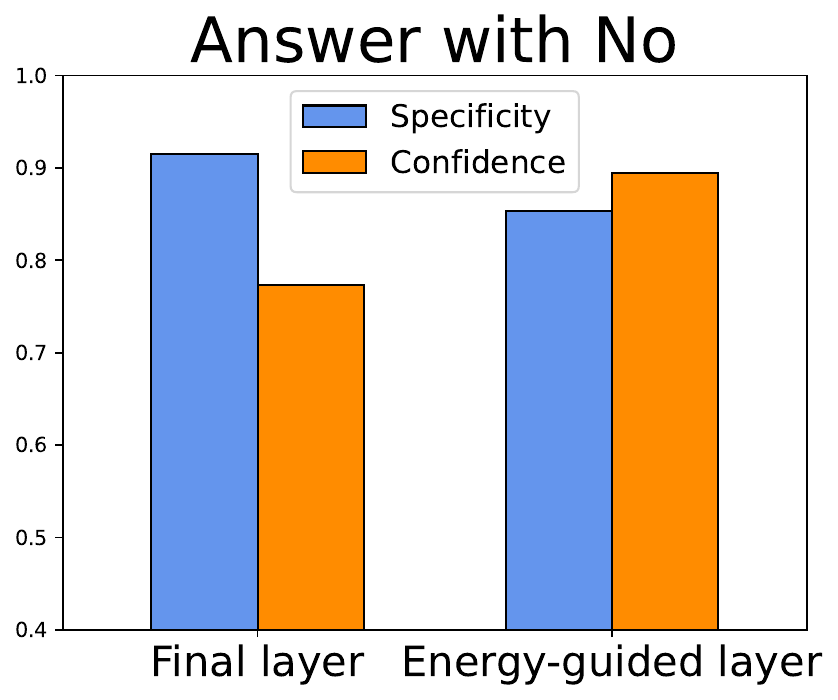}
    \caption{\emph{Accuracy vs. Confidence} for answers with yes (left) and the ones with no (right), using hidden states from the last layer and the ones with minimum energy for decoding, respectively.  The GQA dataset with \emph{adversarial} setting is utilized along with \emph{greedy} decoding. LLaVA-1.5~\cite{llava_improved} is utilized the VLM backbone. }
    \label{acc_conf}
   
\end{figure}

\paragraph{Accuracy vs. confidence}  We visualize the accuracy and confidence for answers with `` Yes '' and ``No'' in Figure~\ref{acc_conf}. The accuracy of answers with ``Yes'' and answers with ``No'' can be calculated as precision and specificity, respectively. The confidence of each answer is measured by the predictive probability of the corresponding token, \ie, the probability distribution after the  $\softmax$ layer. Therefore, the confidence shown in Figure~\ref{acc_conf} is the average confidence of answers with ``Yes'' and ``No'', respectively. One can see that the gap between precision and averaged confidence of answers with ``Yes'' is reduced after dynamically selecting the layer based on the corresponding energy score. Similarly, the gap between specificity and averaged confidence of answers with ``No'' is also reduced after applying energy-guided decoding. It indicates that our method (energy-guided layer) provides a better calibrated answer compared to the final layer. Comparisons for other datasets and models can be founded in supplementary material.

\begin{figure}[ht!]
    \centering
    \includegraphics[width=0.68\linewidth]{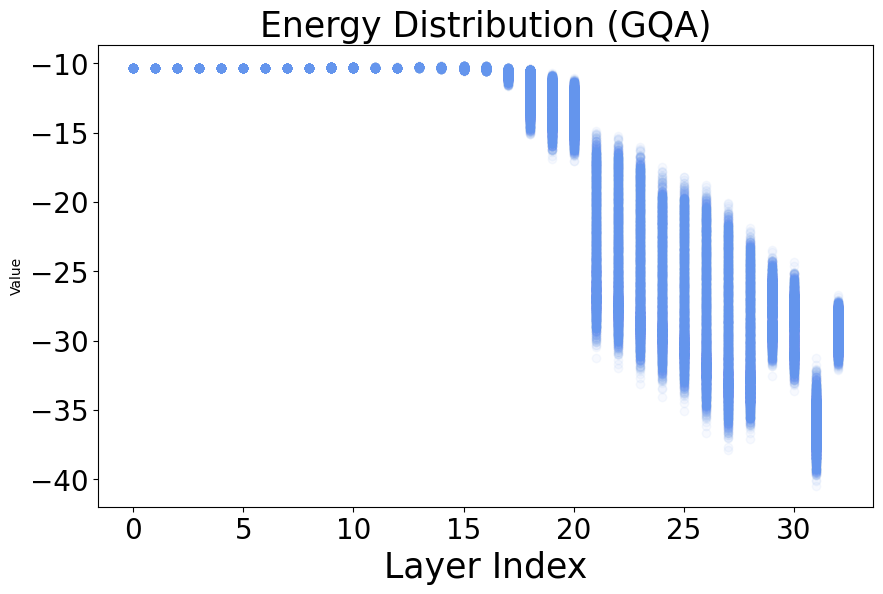}
     
    \caption{\emph{Energy Score Distribution} with LLaVa-1.5~\cite{llava_improved} as the VLM backbone, and \emph{adversarial} setting is utilized for evaluation. }
    \label{energy_score_distribution}
    
\end{figure}

 \paragraph{Energy score visualization} We empirically observe the hidden states selected by the energy-guided decoding mostly come from the second last layer. Therefore, we visualize the energy score across each layer in Figure~\ref{energy_score_distribution} with LLaVA-1.5 as the VLM backbone. One can see the energy score calculated from the penultimate layer is generally lower than other layers, indicating that the corresponding hidden states is more reliable than that from other layers. Visualizations for other datasets and models can be found in the supplementary material.

\section{Conclusion and Discussion}
In this work, we empirically observe a notable bias in terms of ``Yes'' ratio within VLMs when utilizing the hidden states from the final layer for decoding. Moreover, the ``Yes'' ratio biases increase when the tasks are becoming more challenging (\eg, from \textit{random} setting to \textit{adversarial} setting). We leverage ``logit lens''~\cite{logit_lens}, which involves projecting the hidden states extracted from each layer to the ``unembedding matrix'' of the language decoder to obtain multiple logit distributions. To mitigate the object hallucination in the subsequent decoding procedure, we propose to utilize the energy score as a metric to identify the most reliable hidden states. The proposed energy-guided decoding is simple and effective, leading to improved performance in terms of accuracy and F1 score, with a reduced yes-ratio gap. While we primarily focus on the ``Yes'' ratio bias in VLMs, which may originate from the language prior~\cite{gavie}. We hypothesize that similar biases and potential issues in LLMs are also expected to transfer to VLMs, as demonstrated by the ``Yes'' ratio bias empirically observed in our study.

{
    \small
    \bibliographystyle{ieeenat_fullname}
    \bibliography{main}
}
\clearpage
 
 \clearpage
 \appendix
\twocolumn[
  \begin{center}
    {\Large \textbf{Supplementary Material}}
  \end{center}
  \vspace{1cm}
]

\section{Experimental results}
In this section, we show detailed hyperparameter settings and results, including accuracy, precision, recall, F1 score, Yes ratio and yes-ratio gap for three baseline methods including \VCD~, \HALC~, and \OPERA~. All experiments are running on an NVIDIA GeForce RTX 3090 GPU, CUDA 11.4 + PyTorch 2.0.0.

\subsection{Datasets}
 We utilize the data list provided by~\VCD~ for the POPE~\cite{pope} benchmark, available at~\url{https://github.com/DAMO-NLP-SG/VCD/tree/master/experiments/data/POPE}. For the \MMVP~ benchmark, we first correct one error in the dataset, \ie, the answers for image-question pairs 279 and 280 are incorrect. Besides, we select the image-question pairs that requires to answer  ``Yes'' or  ``No'' and report the same metrics as for the POPE~\cite{pope} benchmark. The template for query VLMs is \textit{Is there a $\{\}$ in the image? } for all benchmarks as conducted in VCD~\cite{vcd}.

\subsection{CHAIR Evaluation}
Caption Hallucination Assessment with Image Relevance (CHAIR) is commonly employed metric for object hallucination in captioning tasks~\cite{objectHallucination}. Two variants of CHAIR, \ie, CHAIR$_I$ for evaluating the degree of hallucination at the object instance level and CHAIR$_S$ for evaluating at the sentence level. It is worthwhile to note that CHAIR$_I$  might not be able to reflect the contextual understanding, while overemphasize on individual instances. Formally,
 
\begin{align}
    \text{CHAIR}_I &= \frac{|\{\text{hallucinated objects}\}|}{|\{\text{all mentioned objects}\}|},   
\end{align}
\begin{align}
    \text{CHAIR}_S &= \frac{|\{\text{captions with hallucinated objects}\}|}{|\{\text{all captions}\}|}.
\end{align}

\subsection{Hyperparameter settings}

We conduct the experiments with the hyperparameter setting implemented in HALC~\cite{halc} to ensure the fair comparison. We mainly focus on the discriminative tasks, i.e.\ only the first word is taken consideration for the evaluation. Therefore, we set the number of maximum tokens to be 16 to enable faster inference. The temperature is set to~1 for all experiments.

\begin{table}[h]
 
\begin{adjustbox}{width=\linewidth,center}
 
\centering
\begin{tabular}{l|ll }
\toprule
\textbf{Decoding methods} & \textbf{Parameters}                & \textbf{Value}  \\ 
\midrule
\multirow{3}{*}{\VCD~} & Amplification Factor  $\alpha$ & 1 \\ \cline{2-3} \noalign{\smallskip}
& Adaptive Plausibility Threshold $\beta$   &   0.1 \\ \cline{2-3} \noalign{\smallskip}
& Noise Step &  500 \\
\midrule
\multirow{5}{*}{\HALC~} & Contrast weight $\alpha$        & 0.05 \\ \cline{2-3} \noalign{\smallskip}
 
& JSD Candidate number $k$                 & 6    \\ \cline{2-3} \noalign{\smallskip}
& Number of Sampled FOVs $n$           & 4 \\ \cline{2-3} \noalign{\smallskip}
& Exponential Growth factor $\lambda$  & 0.6 \\ \cline{2-3} \noalign{\smallskip}
& Adaptive Plausibility Threshold $\beta$     & 0.1 \\ 
\midrule
\multirow{4}{*}{\OPERA~} & Self-attention Weights Scale Factor $\theta$ & 50 \\ \cline{2-3} \noalign{\smallskip}
& Attending Retrospection Threshold    &   15 \\ \cline{2-3} \noalign{\smallskip}
& Beam Size &  3 \\\cline{2-3} \noalign{\smallskip}
& Penalty weights & 1 \\
\midrule

 \multirow{4}{*}{\PAI~} & Step Size for Attention Intervention $\alpha$ & 0.2     \\ \cline{2-3} \noalign{\smallskip}
& Degree of Penalty Applied to the Initial
Prediction Distribution $\gamma$   &   1.1 \\ \cline{2-3} \noalign{\smallskip}
& Start layer &  2 \\\cline{2-3} \noalign{\smallskip}
& End layer & 32 \\
\midrule
\end{tabular}
\end{adjustbox}
\caption{\emph{Hyperparameter settings} for the baseline methods.}
\label{tab:hyper-specification}
\end{table}

\subsection{Detailed results on POPE benchmark}
The results for hallucination mitigation on the POPE~\cite{pope} benchmark with LLaVA-1.5~\cite{llava_improved} and InstructBLIP~\cite{instructblip} as the vision-language model (VLM) backbone are presented in Table~\ref{full_results_llava} and Table~\ref{full_results_instructionblip}, respectively. We also include results for \OPERA~ that necessitates the computationally costly beam search decoding. 
One can see that energy-guided decoding (our method) consistently obtains the best results in terms of accuracy, F1 score, and yes-ratio gap across two datasets including A-OKVQA~\cite{a-okvqa} and GQA~\cite{gqa} with three different configurations including \textit{random}, \textit{popular}, and \textit{adversarial} when utilizing LlaVA-1.5~\cite{llava_improved} as the VLM backbone. Specifically, the average accuracy improvement is $4.37\%$  and the average yes-ratio gap reduction is 8.11$\%$ compared to vanilla greedy decoding. \OPERA~, as one of the competitive baseline method, is inferior to our method on GQA dataset across three settings in terms of accuracy, F1 score, and yes-ratio gap. Particularly, our method outperforms \OPERA~ with a margin $1.37\%$ and $3.97\%$ in terms of accuracy and yes-ratio gap, respectively.  The results with InstructBLIP~\cite{instructblip} as the VLM backbone follow a similar, but less pronounced pattern.

\subsection{Detailed results on the MMVP benchmark}
The results of hallucination mitigation on the subset of the MMVP~\cite{mmvp} benchmark with LLaVA-1.5~\cite{llava_improved} and InstructBLIP~\cite{instructblip} as the vision-language model (VLM) backbone are presented in Table~\ref{tab:mmvp-subset-full}. We also include the results in terms of yes ratio. One can see that, energy-guided decoding (our method) consistently obtains the best accuracy and yes-ratio gap across  two architectures. Specifically, the average yes-ratio gap is reduced by a margin of 15.59$\%$ compared to the greedy decoding. 
 
\begin{table}[h]
\begin{adjustbox}{width=\linewidth,center}
\begin{tabular}{@{}llllllll@{}}
\toprule
 Model    & \textbf{Decoding}  & Accuracy$\uparrow$ & Precision & Recall & F1 Score$\uparrow$  &  Yes ratio & $\Delta_\text{gap} \downarrow$ \\ 
 \toprule
 \multirow{4}{*}{LLaVA-1.5}  
 & Regular &59.02&55.91&85.25&67.53& 76.23& \second{26.23}\\
  & Greedy &57.38&54.46&90.16&67.90& 82.79 &32.79\\
 & \VCD~ &\second{60.66}&56.19&96.72& \best{71.08}& 86.07 & 36.07 \\ 
  & \textbf{Energy (Ours)}  & \best{64.75} &  62.50 &  73.77 & \second{67.67} & 59.02 & \best{9.02} \\
\midrule
 \multirow{4}{*}{InstructBLIP}   
  & Regular & 55.74 & 55.07 & 62.30 & 58.46 & 56.55&  \second{6.55} \\
  & Greedy & \second{63.93} & 61.04 & 77.09& \best{68.12} & 63.15 & 13.15  \\
  & \VCD~ &52.46&52.05&62.30&56.72 & 59.84 &9.84                                          \\
  & \textbf{Energy (Ours)}  & \best{64.75} & 63.24 & 70.49 &  \second{66.67} & 55.74 & \best{5.74}  \\
\bottomrule
\end{tabular}
\end{adjustbox} 
\caption{\emph{Results on MMVP dataset.}  
Higher accuracy and F1 score indicate better performance and fewer hallucinations. Lower yes-ratio gap ($\Delta_\text{gap}$) 
implies the model is better calibrated.  
The best entries within each setting are in \textbf{bold}, the 2nd is \underline{underlined}. }
\label{tab:mmvp-subset-full}
\end{table}

\subsection{Yes ratio transfer under nucleus sampling}
\begin{figure}[tbh]
    \centering
    \includegraphics[width=0.32\linewidth]{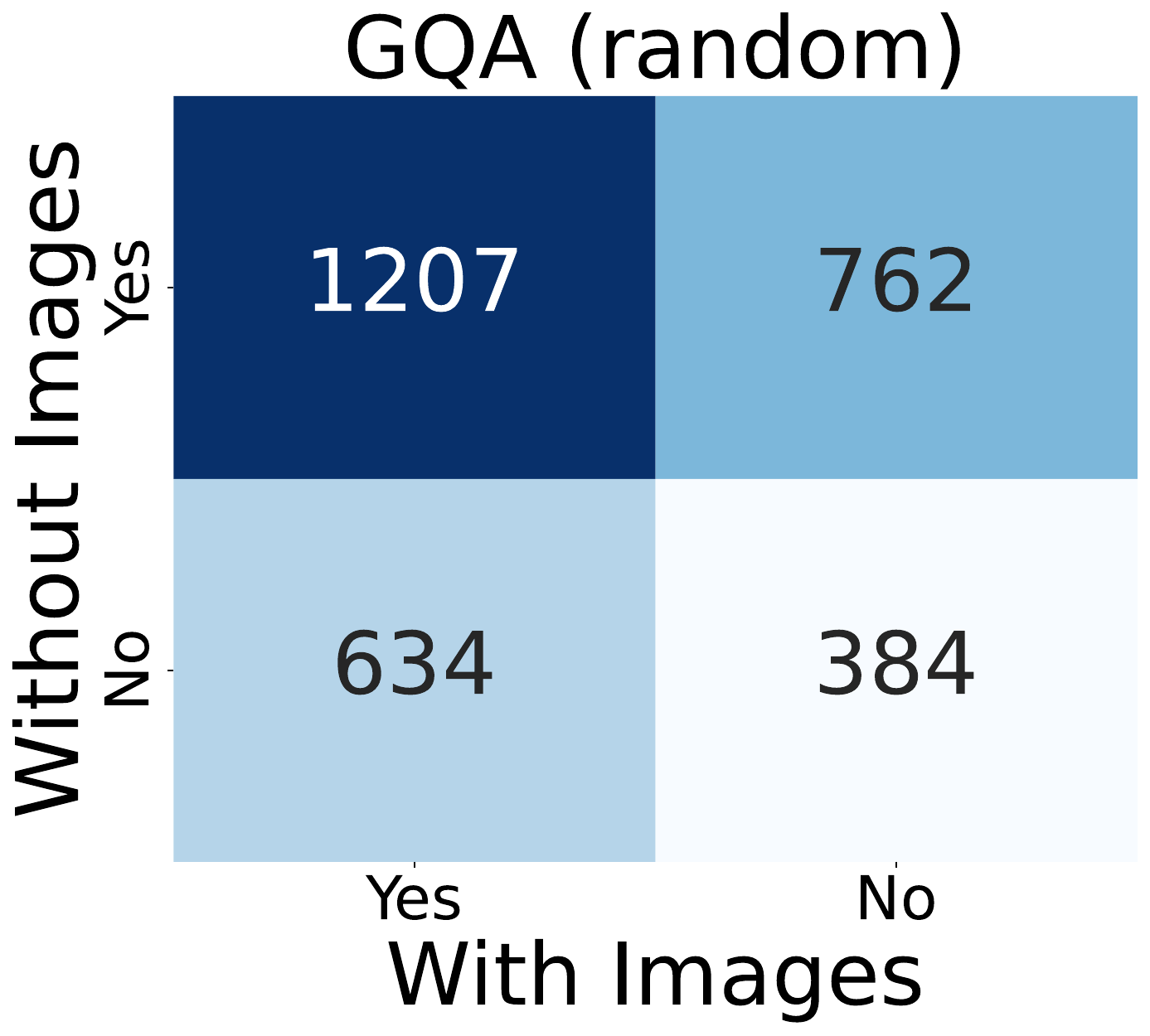}
     \includegraphics[width=0.32\linewidth]{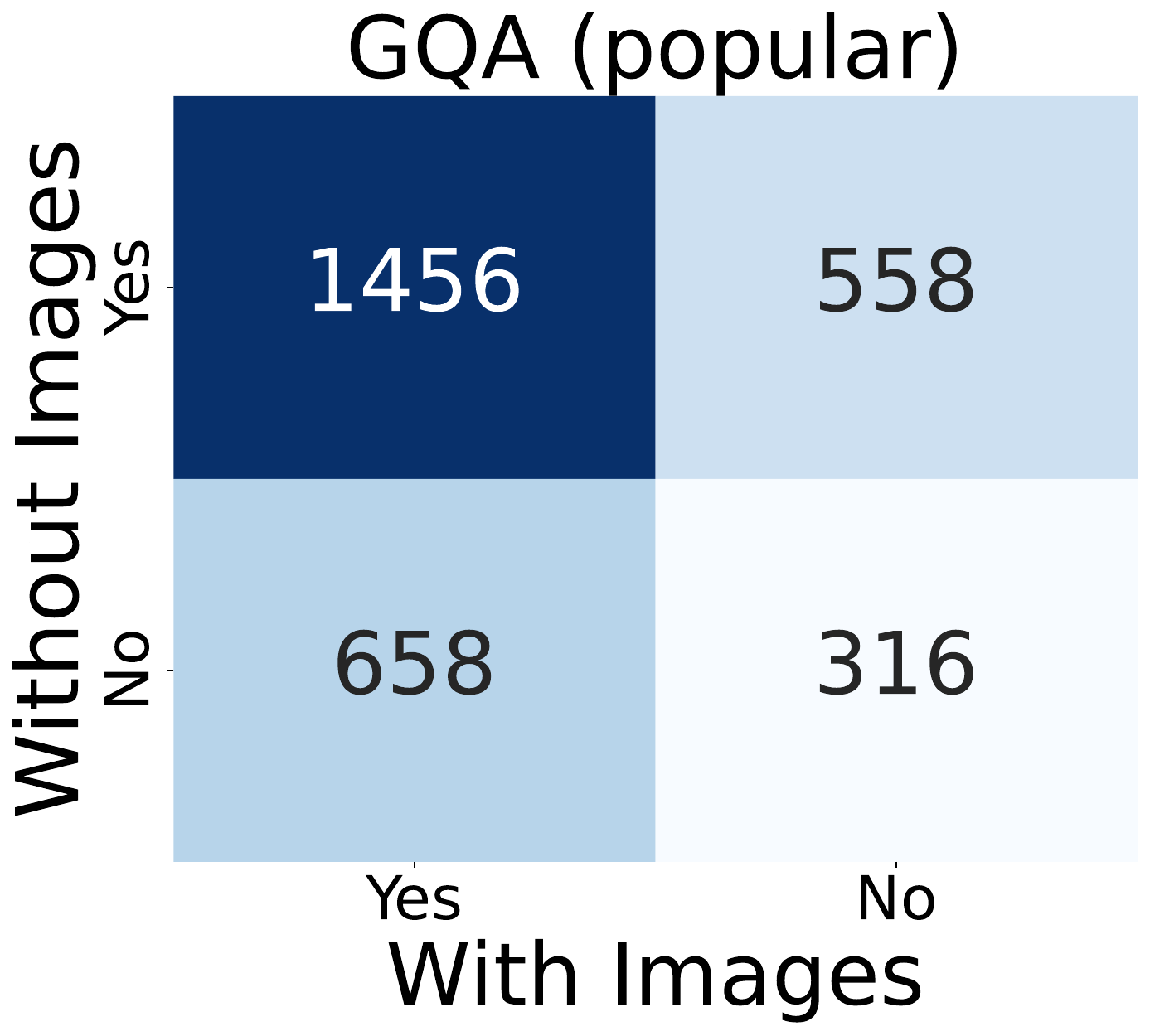}
      \includegraphics[width=0.32\linewidth]{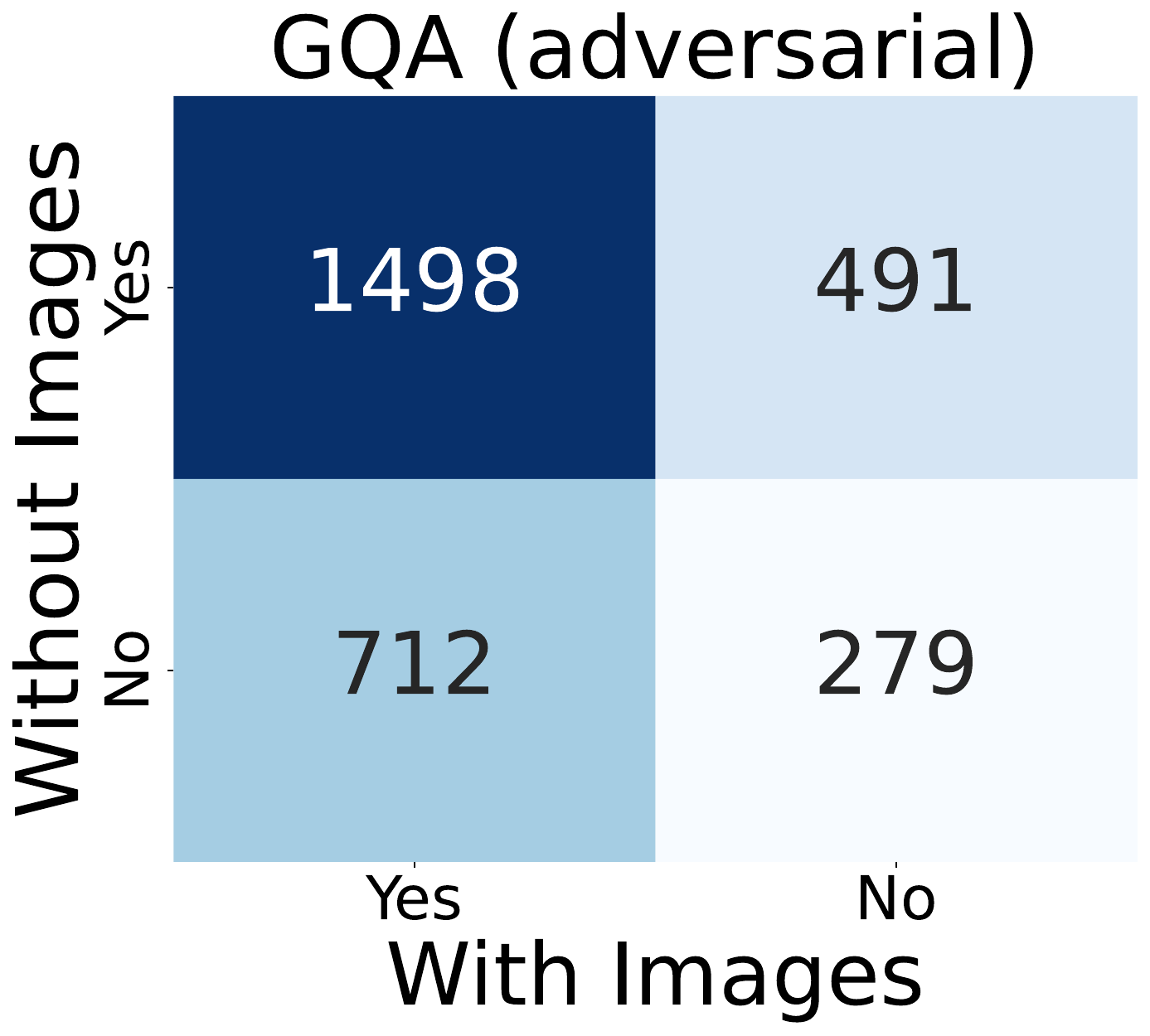}
    \caption{\emph{Transfer of ``Yes'' ratio} from non-visual input to visual inputs using nucleus sampling. Three settings of POPE-GQA~\cite{pope} are utilized including random (left column), popular (middle column), and adversarial (right column). LLaVA-1.5~\cite{llava_improved} is employed as the VLM backbone.}
    \label{yes_ratio_direct}
\end{figure}

We study the yes ratio transfer when using nucleus sampling. The experimental setting is similar to the ones using greedy sampling. Specifically,  the evaluation covering two scenarios—one that includes visual input and one without visual input and the results are shown in Fig.~\ref{yes_ratio_direct}. Each number in the ``confusion matrix'' represents the number of samples (image-question pairs) that overlap between cases with and without visual inputs. For instance, 1207 in the left plot represents the number of image-question pairs that consistently generate the answer `` Yes'' regardless of whether visual inputs are provided. One can see from Fig.~\ref{yes_ratio_direct} that the VLM exhibits a similar pattern as using greedy decoding, i.e., the model tends to answer ``Yes'' when the VQA tasks are becoming more difficult (from \emph{random} to \emph{adversarial}) and  the ``Yes'' ratio is initially high without visual input, suggesting models are biased toward ``Yes'' due to language priors in this dataset.

 \begin{figure}[h]
    \centering
    \includegraphics[width=0.7\linewidth]{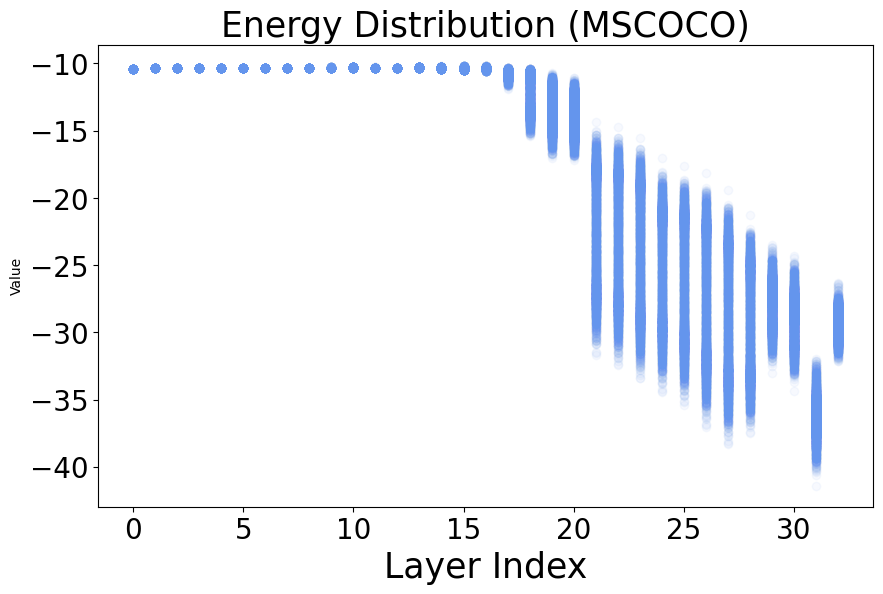}
    \includegraphics[width=0.7\linewidth]{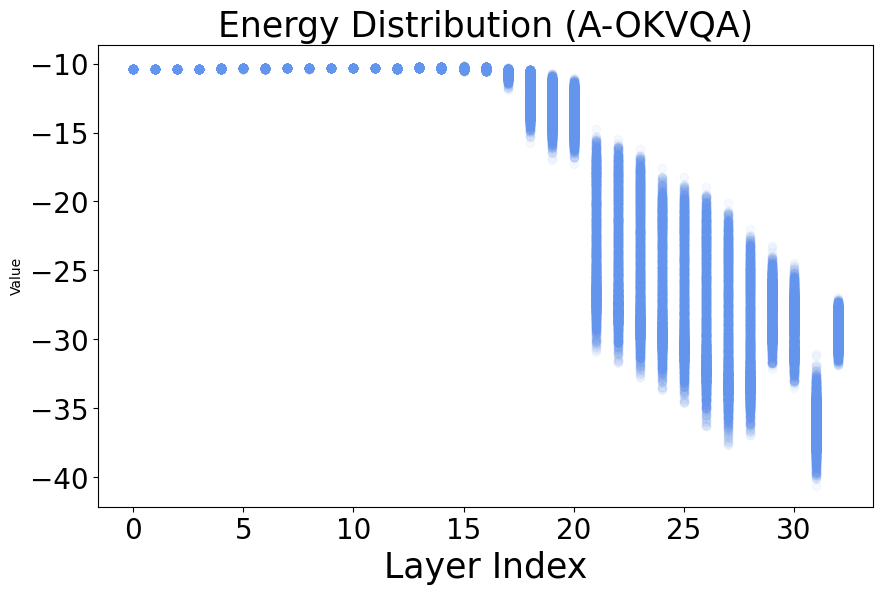}
    \includegraphics[width=0.7\linewidth]{sec/figures/gqa_energy_visualization.png}
    \caption{\emph{Energy distribution} at each layer with LLaVa-1.5~\cite{llava_improved} as the VLM backbone. Three datasets including MSCOCO~\cite{mscoco}, A-OKVQA~\cite{a-okvqa}, and GQA~\cite{gqa} with \emph{adversarial} setting are utilized along with \emph{greedy} decoding. LLaVA-1.5~\cite{llava_improved} is utilized as the VLM backbone.}
    \label{energy_score_distribution_full}
\end{figure}

\begin{table*}[ht!]
\begin{adjustbox}{width=.75\linewidth,center}
\begin{tabular}{cllllllll}
\toprule
\textbf{Dataset}          & \textbf{Setting}                         & \textbf{Decoding} & Accuracy$\uparrow$ & Precision & Recall & F1 Score$\uparrow$ & Yes ratio  & $\Delta_\text{gap} \downarrow$ \\ \midrule
\toprule

\multirow{15}{*}{\rotatebox{90}{MSCOCO}}      & \multirow{5}{*}{\textit{Random}}      & Greedy &\best{89.37}&89.66&89.00&\best{89.33} &  49.63 & \best{0.37}                       \\
                          &                                                             & \HALC~ &\second{89.30}&89.70&88.80&\second{89.25}&49.5  & \second{0.5}                        \\
                          &                                                             & \VCD~  &84.83&82.76&88.00&85.30&53.17  & 3.17                      \\
                          &                                                              & \OPERA~ &89.17&92.48&85.27&88.73&46.10 & 3.90              \\ 
                          &                                                              &\PAI~ &89.30&89.54&89.01&89.27&49.72&0.28 \\
                          &                                                             &  \textbf{Energy (Ours)} &87.50&96.07&78.20&86.22&40.70 & 9.30   \\ \cline{2-9} \noalign{\smallskip}

                          & \multirow{5}{*}{\textit{Popular}}                           & Greedy &86.00&83.96&89.00&86.41&53.00  & 3.00        \\
                          &                                                             & \HALC~  & \second{86.10}&84.25&88.80&\second{86.47}&52.70 &\second{2.7}                      \\
                          &                                                             & \VCD~ &81.77&78.31&87.87&82.81&56.10 &6.10                    \\
                          &                                                             & \OPERA~ &\best{86.80}&87.96&85.27&\best{86.59}&48.47 & \best{1.53}\\
                          &                                                              & \PAI~&86.20&84.28&89.00&86.58&52.80&2.80\\
                          &                                                             &  \textbf{Energy (Ours)} &85.80&92.22&78.20&84.63&42.4 & 7.6 \\ \cline{2-9} \noalign{\smallskip}  
                          
                          & \multirow{5}{*}{\textit{Adversarial}}                       & Greedy  &79.10&74.34&88.87&80.96&59.77 & 9.77      \\
                          &                                                             & \HALC~  &79.27&74.64&88.67&81.05&59.40 & 9.40                       \\
                          &                                                             & \VCD~  &76.17&71.09&88.20&78.73&62.03 & 12.03                      \\
                          &                                                             & \OPERA~ &\second{81.20}&78.89&85.20&\second{81.92}&54.00 & \second{6.00}\\
                          &                                                              &\PAI~&79.30&74.59&88.87&81.11&59.57&9.57\\
                          &                                                             &  \textbf{Energy (Ours)} &\best{82.90}&86.42&78.07&\best{82.03}& 45.17 & \best{4.83}\\  \hline

\multirow{15}{*}{\rotatebox{90}{A-OKVQA}}      & \multirow{5}{*}{\textit{Random}}        & Greedy &85.70&80.17&94.87&86.90&59.17 & 9.17                    \\
                          &                                                             & \HALC~&85.80&80.30&94.87&86.98&59.07 & 9.07                       \\
                          &                                                             & \VCD~ &80.77&74.75&92.93&82.85&62.17  & 12.17                   \\
                          &                                                             & \OPERA~ &\second{88.23}&86.09&91.20&\best{88.57}&52.97  & \second{2.97}\\
                          &                                                             &\PAI~&85.90&80.44&94.87&87.06&58.97&8.97 \\
                          &                                                             &  \textbf{Energy (Ours)} &\best{88.60}&90.72&86.00&\second{88.30}& 47.4 & \best{2.6} \\ \cline{2-9} \noalign{\smallskip}

                          & \multirow{5}{*}{\textit{Popular}}                           & Greedy &79.90&73.01&94.87&82.52&64.97 & 14.97        \\
                          &                                                             & \HALC~ &79.97&73.09&94.87&\second{82.56}& 64.9 & 14.90                   \\
                          &                                                             & \VCD~ &76.47&69.85&93.13&79.83&66.67  & 16.67                     \\
                          &                                                             & \OPERA~  &\second{83.37}&78.85&91.20&84.57&57.83 & \second{7.83}\\
                          &                                                             &\PAI~&80.33&73.50&94.87&82.83&64.53&14.53\\
                          &                                                             &  \textbf{Energy (Ours)} &\best{84.67}&83.77&86.00&\best{84.87}& 51.33 & \best{1.33} \\  \cline{2-9} \noalign{\smallskip}  
                         
                          & \multirow{5}{*}{\textit{Adversarial}}                       & Greedy&69.07&62.58&94.87&75.41&75.80 & 25.8         \\
                          &                                                             & \HALC~  &69.23&62.71&94.87&75.51&75.63  & 25.63                     \\
                          &                                                             & \VCD~  &68.47&62.50&92.33&74.54&73.87 & 23.87                     \\                
                          &                                                             & \OPERA~ &\second{73.90}&67.76&91.20&\second{77.75}&67.30   & \second{17.30}               \\  
                          &                                                             &\PAI~&69.60&63.02&94.87&75.73&75.27&25.27\\
                          &                                                             &  \textbf{Energy (Ours)} & \best{77.40}&73.38&86.00&\best{79.19}&58.59& \best{8.59}   \\ \hline

\multirow{15}{*}{\rotatebox{90}{GQA}}      & \multirow{5}{*}{\textit{Random}}           & Greedy &85.77&79.69&96.00&87.09&60.23 & 10.23                     \\
                          &                                                             & \HALC~&85.90&79.87&96.00&87.19&60.10 & 10.10                    \\
                          &                                                             & \VCD~ &81.33&75.24&93.40&83.34&62.07 & 12.07                 \\
                          &                                                             & \OPERA~ &\second{88.57}&85.47&92.93&\second{89.05}&54.37 & \second{4.37} \\
                          &                                                             &\PAI~&86.13&80.14&96.07&87.39&59.93&9.93\\
                          &                                                             &  \textbf{Energy (Ours)}&\best{89.37}&90.70&87.73&\best{89.19}& 48.37 & \best{1.63}  \\ \cline{2-9} \noalign{\smallskip} 
 
                          & \multirow{5}{*}{\textit{Popular}}                           & Greedy &74.73&67.35&96.00&79.16 & 71.27 & 21.27       \\
                          &                                                             & \HALC~  &74.87&67.48&96.00&79.25&71.13   & \second{21.13}                \\
                          &                                                             & \VCD~ &71.53&64.79&94.33&76.82&72.8 & 22.80                   \\
                          &                                                             & \OPERA~ &\second{79.83}&73.64&92.93&\second{82.17}&63.10 & 23.10     \\
                          &                                                             &\PAI~&75.27&67.84&96.07&79.53&70.80&20.80\\
                          &                                                             &   \textbf{Energy (Ours)}&\best{82.53}&79.47&87.73&\best{83.40}&55.20 & \best{5.20}\\ \cline{2-9} \noalign{\smallskip}  
                          & \multirow{5}{*}{\textit{Adversarial}}                       & Greedy&69.43&62.69&96.00&75.85 & 76.57 & 26.57         \\
                          &                                                             & \HALC~  &69.53&62.77&96.00&75.91&76.47  & 26.47                \\
                          &                                                             & \VCD~  &68.97&62.67&93.80&75.14&74.83 &  24.83                 \\ 
                          &                                                              & \OPERA~ &\second{75.00}&68.40&92.93&\second{78.80}&67.93& \second{17.93}\\
                          &                                                              &\PAI~&70.00&63.15&96.07&76.20&76.07&26.07\\
                          &                                                             &   \textbf{Energy (Ours)}&\best{79.63}&75.50&87.73&\best{81.16}& 58.09 & \best{8.09} \\
                                                                                                                 \hline
\bottomrule
\end{tabular}
\end{adjustbox} 
\caption{ \emph{Results on POPE benchmark with \textbf{LLaVA-1.5}~\cite{llava_improved} as the model.} Higher accuracy and F1 score indicate better performance and fewer hallucinations. Lower yes-ratio gap, ($\Delta_\text{gap}$ )
implies the model is better calibrated.The best performing method within each setting in \textbf{bold}.}
\label{full_results_llava}
\vspace{+2em}
\end{table*}


\begin{table*}[ht!]
\begin{adjustbox}{width=.85\linewidth,center}
\begin{tabular}{cllllllll}
\hline
\textbf{Dataset}          & \textbf{Setting}                         & \textbf{Decoding} & Accuracy$\uparrow$ & Precision & Recall & F1 Score$\uparrow$ & Yes ratio & $\Delta_\text{gap} \downarrow$   \\ \midrule
\toprule

\multirow{15}{*}{\rotatebox{90}{MSCOCO}}      & \multirow{5}{*}{\textit{Random}}      & Greedy &\best{90.17}&92.76&87.13&\best{89.86}&46.97 & 3.03                     \\
                          &                                                             & \VCD~ &84.47&84.84&83.93&84.38& 49.47 & \best{0.03}                       \\
                          &                                                             & \HALC~ &89.73&91.45&87.67&\second{89.52}&47.93  & \second{2.07}                     \\
                          &                                                              & \OPERA~ &\second{89.83}&93.71&85.40&89.36&45.57 & 4.43               \\           
                          &                                                             &  \textbf{Energy (Ours)} &86.80&97.67&75.40&85.10&38.60 & 11.40 \\ \cline{2-9} \noalign{\smallskip}

                          & \multirow{5}{*}{\textit{Popular}}                           & Greedy &83.47&81.18&87.13&\second{84.05}&53.67 & \second{3.67}       \\
                          &                                                             & \VCD~   &77.73&74.47&84.40&79.12&56.67 & 6.67                     \\
                          &                                                             & \HALC~  &82.30&79.17&87.67&83.20&55.37  & 5.37                 \\
                          &                                                             & \OPERA~  &\best{84.67}&84.17&85.40&\best{84.78}& 50.73 & \best{0.73} \\
                          &                                                             &  \textbf{Energy (Ours)} &\second{83.70}&90.41&75.40&82.22&41.70 & 8.30\\ \cline{2-9} \noalign{\smallskip}  
                         
                          & \multirow{5}{*}{\textit{Adversarial}}                       & Greedy &\second{80.67}&77.22&87.00&\second{81.82}&56.33 & \second{6.33}        \\
                          &                                                             & \VCD~ &75.87&71.77&85.27&77.94&59.40 & 9.40                       \\
                          &                                                             & \HALC~  &79.47&75.40&87.47&80.99&58.00  & 8.00                      \\
                          &                                                             & \OPERA~ &81.43&79.20&85.27&\best{82.12}& 53.83 & \best{3.83} \\
                          &                                                             &  \textbf{Energy (Ours)} &\best{82.17}&87.09&75.53&80.90&43.37 & 6.63\\  \hline

\multirow{15}{*}{\rotatebox{90}{A-OKVQA}}      & \multirow{4}{*}{\textit{Random}}        & Greedy &\second{89.13}&86.60&92.60&\best{89.50}&53.47 & \second{3.47}                   \\
                          &                                                             & \VCD~ &83.23&79.51&89.53&84.23&56.30 & 6.30                     \\
                          &                                                             &  \HALC~  &88.27&84.66&93.47&88.85&55.20 & 5.20                   \\
                          &                                                             & \OPERA~  &\best{89.57}&88.97&90.33&\second{89.65}&50.77 & \best{0.77}  \\
                          &                                                             &  \textbf{Energy (Ours)} &89.07&93.80&83.67&88.44&44.60 & 5.40\\ \cline{2-9} \noalign{\smallskip}

                          & \multirow{5}{*}{\textit{Popular}}                           & Greedy &79.57&73.45&92.60&81.92&63.03 & 13.03         \\
                          &                                                             & \VCD~  &76.87&70.95&91.00&79.73&64.13 & 14.13                  \\
                          &                                                             &  \HALC~ &78.20&71.60&93.47&81.09&65.27 & 15.27                  \\
                          &                                                             & \OPERA~ &\second{82.67}&78.32&90.33&\second{83.90}&57.67  & \second{7.67} \\
                          &                                                             &  \textbf{Energy (Ours)} &\best{84.03}&84.28&83.67&\best{83.97}&49.63 & \best{0.37}\\  \cline{2-9} \noalign{\smallskip}  
                         
                          & \multirow{5}{*}{\textit{Adversarial}}                       & Greedy&71.43&65.06&92.60&76.42&71.17& 21.17         \\
                          &                                                             & \VCD~ &69.23&63.81&88.87&74.28&69.63  & 19.63                    \\
                          &                                                             & \HALC~  &70.33&63.90&93.47&75.91&73.13  & 23.13                  \\                &                                                                            & \OPERA~   &\second{74.13}&68.23&90.33&\second{77.74}&66.20  & \second{16.20}              \\                           
                          &                                                             &  \textbf{Energy (Ours)}  &\best{76.70}&73.43&83.67&\best{78.22}&56.97 & \best{6.97}  \\ \hline

\multirow{15}{*}{\rotatebox{90}{GQA}}      & \multirow{4}{*}{\textit{Random}}           & Greedy  &\second{86.90}&84.70&90.07&\second{87.30}&53.17 & \second{3.17}                     \\
                          &                                                             & \VCD~ &80.90&77.35&87.40&82.07&56.50 & 6.50                     \\
                          &                                                             &  \HALC~   &85.97&83.08&90.33&86.55&54.37 & 4.37                \\
                          &                                                             & \OPERA~  &\best{87.33}&87.23&87.47&\best{87.35}&50.13 & \best{0.13} \\
                          &                                                             &  \textbf{Energy (Ours)}  &86.53&92.35&79.67&85.54&43.13 & 6.87\\ \cline{2-9} \noalign{\smallskip} 
 
                          & \multirow{5}{*}{\textit{Popular}}                           & Greedy&76.37&70.70&90.07&79.21&63.70 & 13.70       \\
                          &                                                             & \VCD~   &73.00&67.97&87.00&76.32&64.00  & 14.00              \\
                          &                                                             & \HALC~   &74.50&68.61&90.33&77.99&65.83 &15.83              \\
                          &                                                             & \OPERA~ &\second{79.77}&75.79&87.47&\best{81.21}&57.70 & \second{7.70}     \\
                          &                                                             &   \textbf{Energy (Ours)} &\best{80.27}&80.63&79.67&\second{80.15}&49.40 & \best{0.60}\\ \cline{2-9} \noalign{\smallskip}  
                          & \multirow{5}{*}{\textit{Adversarial}}                       & Greedy  &71.50&65.68&90.07&75.96&68.57 & 18.57        \\
                          &                                                             & \VCD~ &69.10&64.01&87.27&73.85&68.17 & 18.17                 \\
                          &                                                             & \HALC~ &69.70&63.95&90.33&74.88&70.63  & 20.63              \\ 
                          &                                                              & \OPERA~  &\second{74.00}&68.91&87.47&\second{77.09}&63.47& \second{13.47}  \\
                          &                                                             &   \textbf{Energy (Ours)}&\best{76.57}&75.02&79.67&\best{77.27}&  53.10 & \best{3.10}\\
                                                                                                                 \hline
\bottomrule
\end{tabular}
\end{adjustbox} 
\caption{ \emph{Results on POPE benchmark with \textbf{InstructBLIP}~\cite{instructblip}.} Higher accuracy and F1 score indicate better performance and fewer hallucinations. Lower yes-ratio gap (
$\Delta_\text{gap}$) implies the model is better calibrated. The best performing method within each setting in \textbf{bold}, the 2nd is \underline{underlined}. }
\label{full_results_instructionblip}
\vspace{+4em}
\end{table*}


\begin{table*}[ht!]
\begin{adjustbox}{width=.85\linewidth,center}
\begin{tabular}{cllllllll}
\hline
\textbf{Dataset}          & \textbf{Setting}                         & \textbf{Decoding} & Accuracy$\uparrow$ & Precision & Recall & F1 Score$\uparrow$ & Yes ratio & $\Delta_\text{gap} \downarrow$   \\ \midrule
\toprule

\multirow{15}{*}{\rotatebox{90}{MSCOCO}}      & \multirow{5}{*}{\textit{Random}}      & Greedy &83.30&77.95&92.87&84.76&59.57&9.57                    \\
                          &                                                             & \VCD~ &80.83&75.45&91.40&82.67&60.57&10.57                      \\
                          &                                                             & \HALC~ &83.57&78.30&92.87&84.96&59.30&9.30                     \\
                          &                                                              & \OPERA~ &\second{86.37}&83.32&90.93&\best{86.96}&54.57&\best{4.57}              \\           
                          &                                                             &  \textbf{Energy (Ours)} &\best{87.73}&95.79&78.93&\second{86.55}&41.20&\second{8.80} \\ \cline{2-9} \noalign{\smallskip}

                          & \multirow{5}{*}{\textit{Popular}}                           & Greedy&77.40&70.93&92.87&80.43&65.47&15.47        \\
                          &                                                             & \VCD~   &75.07&68.89&91.40&78.57&66.33&16.33                    \\
                          &                                                             & \HALC~   &77.57&71.11&92.87&80.54&65.30&15.30                     \\
                          &                                                             & \OPERA~&\second{81.37}&76.33&90.93&\second{82.99}&59.57&\second{9.5}7    \\
                          &                                                             &  \textbf{Energy (Ours)} &\best{86.50}&93.01&78.93&\best{85.39}&42.43&\best{7.57}\\ \cline{2-9} \noalign{\smallskip}  
                         
                          & \multirow{5}{*}{\textit{Adversarial}}                       & Greedy &73.80&67.25&92.80&77.98&69.00&19.00        \\
                          &                                                             & \VCD~  &72.80&66.68&91.13&77.01&68.33&18.33                      \\
                          &                                                             & \HALC~ &74.00&67.44&92.80&78.11&68.80&18.80               \\
                          &                                                             & \OPERA~ &\second{77.13}&71.26&90.93&\second{79.91}&63.80&\second{13.80}  \\
                          &                                                             &  \textbf{Energy (Ours)}&\best{84.50}&88.94&78.80&\best{83.56}&44.30&\best{5.70}  \\  \hline

\multirow{15}{*}{\rotatebox{90}{A-OKVQA}}      & \multirow{4}{*}{\textit{Random}}        & Greedy  &79.23&71.40&97.53&82.45&68.30&18.30                 \\
                          &                                                             & \VCD~ &77.50&70.21&95.53&80.94&68.03&18.03                   \\
                          &                                                             &  \HALC~ &79.40&71.60&97.47&82.55&68.07&18.07               \\
                          &                                                             & \OPERA~ &83.80&77.58&95.07&85.44&61.27&11.27    \\
                          &                                                             &  \textbf{Energy (Ours)} &\best{87.97}&91.97&83.20&\best{87.36}&45.23&4.77                \\ \cline{2-9} \noalign{\smallskip}

                          & \multirow{5}{*}{\textit{Popular}}                           & Greedy &71.83&64.42&97.53&77.59&75.70&25.70         \\
                          &                                                             & \VCD~ &71.40&64.42&95.60&76.97&74.20&24.20                    \\
                          &                                                             &  \HALC~ &72.17&64.72&97.47&77.79&75.30&25.30                  \\
                          &                                                             & \OPERA~ &\second{77.93}&70.80&95.07&\second{81.16}&67.13&\second{17.13}                                                    \\
                          &                                                             &  \textbf{Energy (Ours)} &\best{83.67}&83.98&83.20&\best{83.59}&49.53&\best{0.47}  \\ \cline{2-9} \noalign{\smallskip}  
                         
                          & \multirow{5}{*}{\textit{Adversarial}}                       & Greedy &64.80&58.94&97.53&73.48&82.73&32.73         \\
                          &                                                             & \VCD~   &65.40&59.62&95.40&73.38&80.00&30.00                   \\
                          &                                                             & \HALC~  &64.90&59.02&97.47&73.52&82.57&32.57             \\                &                                                            & \OPERA~&\second{69.57}&62.96&95.07&\best{75.75}&75.50&\second{25.50}                 \\                           
                          &                                                             &  \textbf{Energy (Ours)} &\best{77.37}&74.51&83.20&\best{78.61}&55.83&\best{5.83}   \\ \hline

\multirow{15}{*}{\rotatebox{90}{GQA}}      & \multirow{4}{*}{\textit{Random}}           & Greedy &83.17&77.02&94.53&84.88&61.37&11.37                   \\
                          &                                                             & \VCD~  &80.77&75.57&90.93&82.54&60.17&10.17               \\
                          &                                                             &  \HALC~ &82.81&76.56&94.46&\second{84.57}&61.54&11.54               \\
                          &                                                             & \OPERA~ &\best{86.07}&83.69&89.60&\best{86.54}&53.53&\best{3.53}  \\
                          &                                                             &  \textbf{Energy (Ours)}  &\second{85.60}&93.77&76.27&84.12&40.67&\second{9.33}   \\ \cline{2-9} \noalign{\smallskip} 
 
                          & \multirow{5}{*}{\textit{Popular}}                           & Greedy &73.77&66.79&94.53&78.28&70.77&20.77      \\
                          &                                                             & \VCD~ &72.03&65.99&90.93&76.48&68.90&18.90              \\
                          &                                                             & \HALC~ &74.13&67.17&94.40&78.49&70.27&20.27              \\
                          &                                                             & \OPERA~  &\second{76.53}&71.04&89.60&\best{79.25}&63.07&13.07    \\
                          &                                                             &   \textbf{Energy (Ours)} &\best{79.80}&82.07&76.27&\second{79.06}&46.47&\best{3.53}  \\ \cline{2-9} \noalign{\smallskip}  
                          & \multirow{5}{*}{\textit{Adversarial}}                       & Greedy&69.77&63.22&94.53&75.77&74.77&24.77          \\
                          &                                                             & \VCD~ &70.03&64.05&91.33&75.30&71.30&21.30                  \\
                          &                                                             & \HALC~  &70.13&63.55&94.40&75.97&74.27&24.27     \\                    &                                                              & \OPERA~ &\second{73.20}&67.47&89.60&\second{76.98}&66.40&\second{16.40}   \\
                          &                                                             &   \textbf{Energy (Ours)}y&\best{78.40}&79.67&76.27&77.93&47.87&\best{2.13}\\
                                                                                                                 \hline
\bottomrule
\end{tabular}
\end{adjustbox} 
\caption{ \emph{Results on POPE benchmark with \textbf{mPLUG-Owl2}~\cite{mplugowl2}}. Higher accuracy and F1 score indicate better performance and fewer hallucinations. Lower yes-ratio gap (
$\Delta_\text{gap}$) implies the model is better calibrated. The best performing method within each setting in \textbf{bold}, the 2nd is \underline{underlined}. }
\label{full_results_instructionblip}
\vspace{+4em}
\end{table*}

\subsection{Accuracy vs. confidence}
\begin{figure*}[t]
     \centering
     \includegraphics[width=0.25\linewidth]{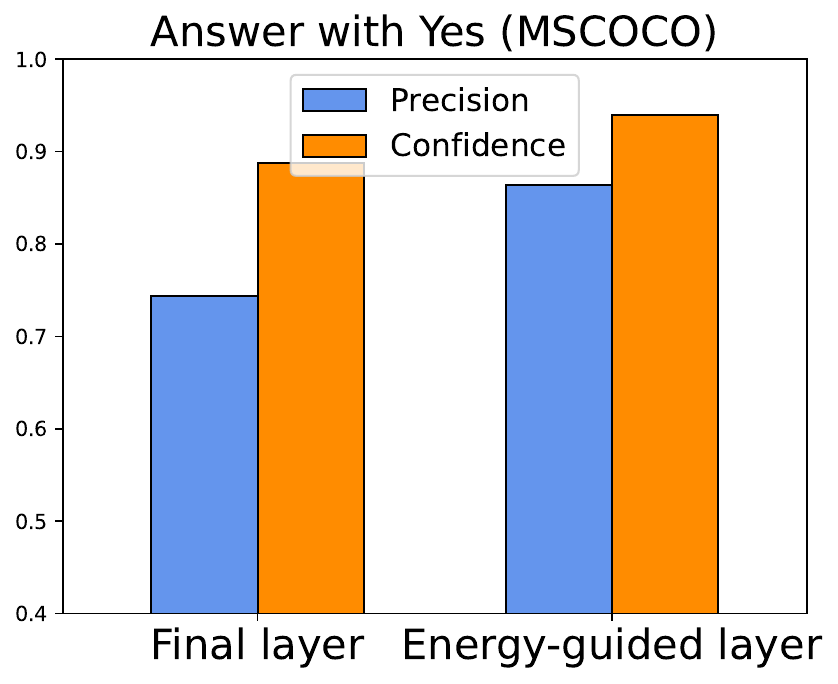}
     \includegraphics[width=0.25\linewidth]{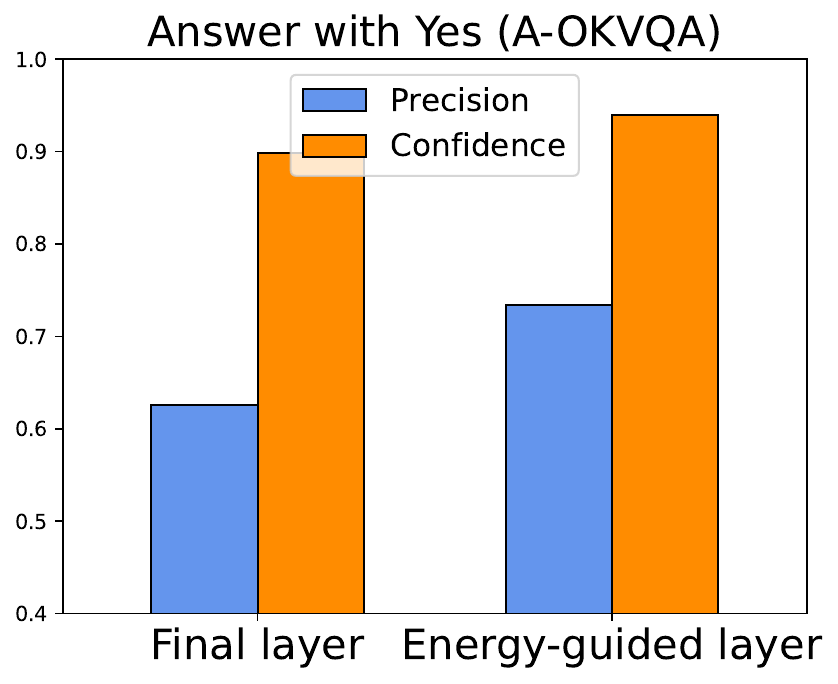}
     \includegraphics[width=0.25\linewidth]{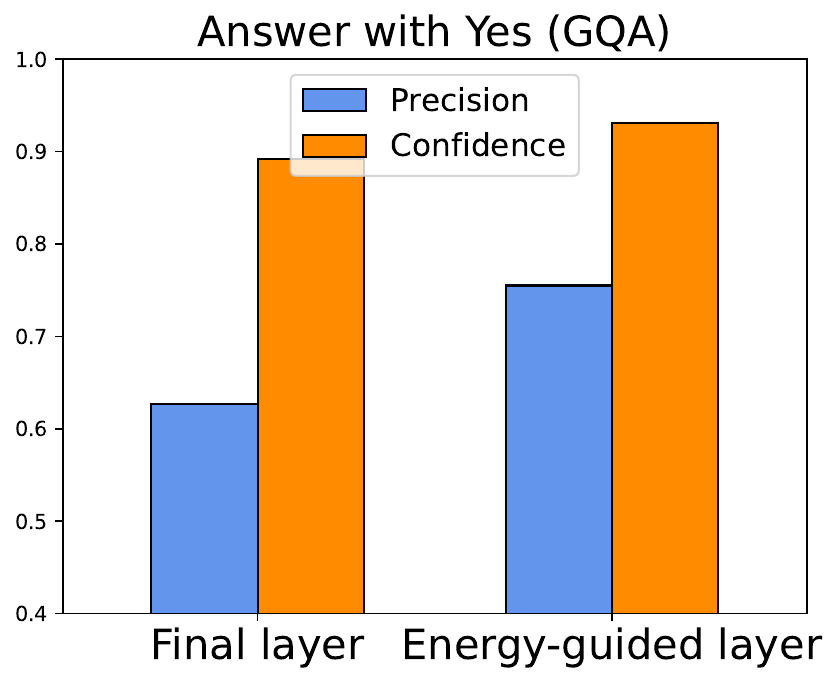}
      \includegraphics[width=0.25\linewidth]{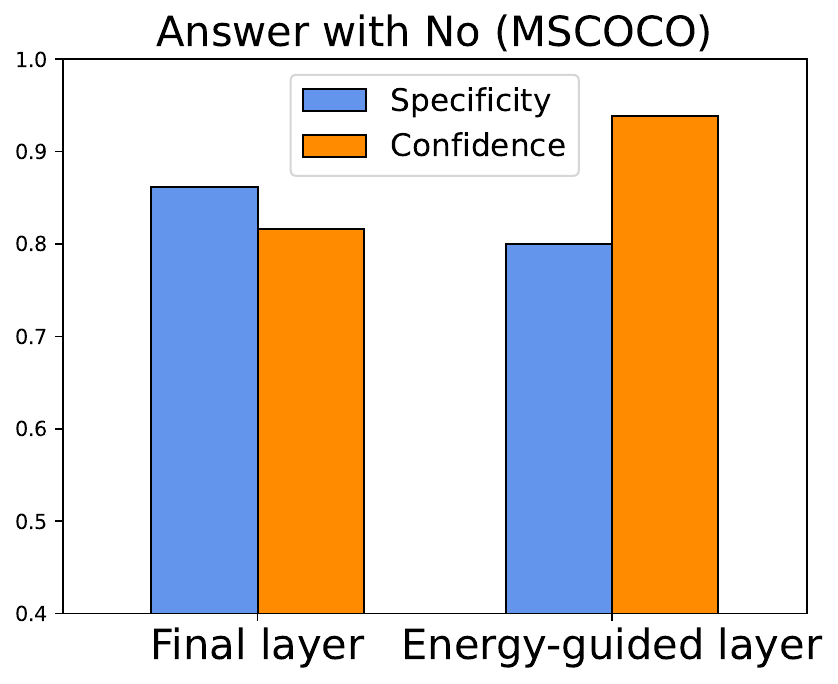}
     \includegraphics[width=0.25\linewidth]{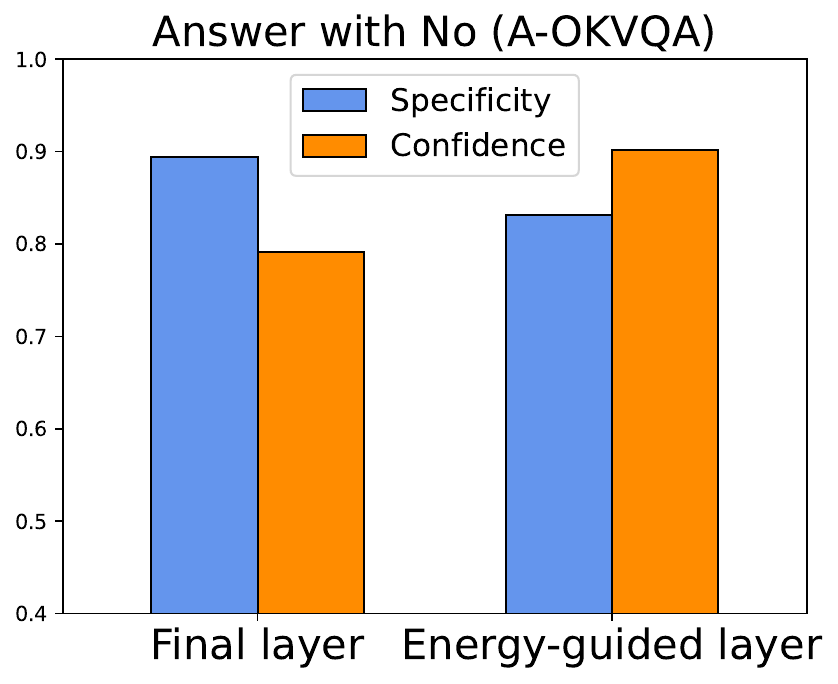}
     \includegraphics[width=0.25\linewidth]{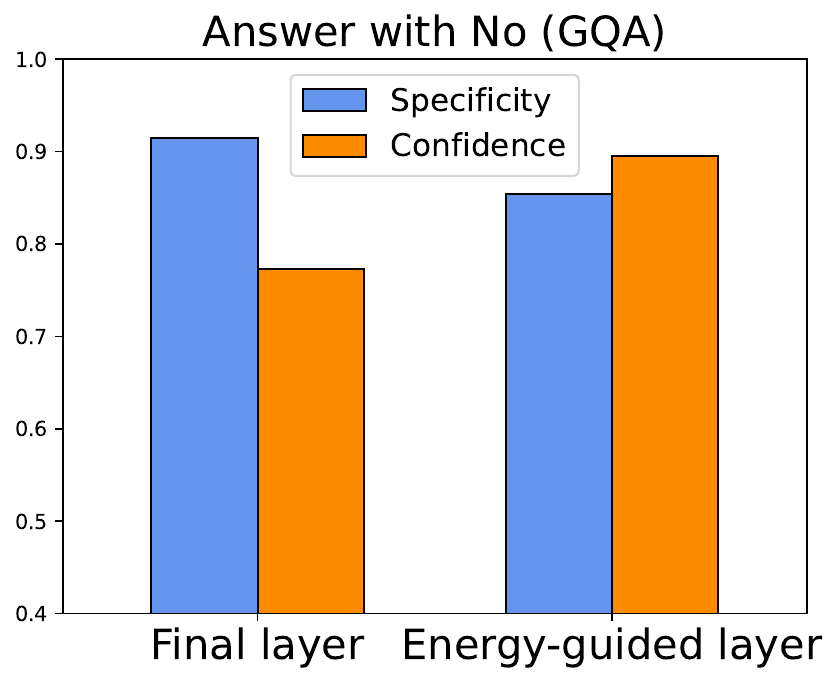}
     \caption{\emph{Accuracy vs. Confidence} for answers with ``Yes'' (top row) and the ones with ``No'' (bottom row), using hidden states from the last layer and the ones with minimum energy for decoding, respectively. Three datasets including MSCOCO, A-OKVQA, and GQA with \emph{adversarial} setting are utilized along with \emph{greedy} decoding. LLaVA-1.5~\cite{llava_improved} is utilized as the VLM backbone. }
     \label{acc_conf_full}
 \end{figure*}

In this section,we visualize the accuracy and confidence for answers with `` Yes '' and `` No'' in Fig.~\ref{acc_conf_full} for three datasets including MSCOCO~\cite{mscoco}, A-OKVQA~\cite{a-okvqa}, and GQA~\cite{gqa} with the \emph{adversarial} setting. The accuracy of answers with ``Yes'' and answers with ``No'' can be calculated as precision and specificity, respectively. The confidence of each answer is measured by the predictive probability of the corresponding token. Additionally, the confidence shown in Fig.~\ref{acc_conf_full} is the average confidence of answers with ``Yes'' and `` No'', respectively. One can see from Fig.~\ref{acc_conf_full} that energy-guided decoding generally narrows the gap between accuracy and confidence, i.e., the gap between precision and confidence for answer with ``Yes'' and the gap between specificity and confidence for answer with ``No''. That is to say, energy-guided decoding provides better calibrated answers.

\subsection{Energy score distribution}
We visualize the energy distribution at every layer in Fig.~\ref{energy_score_distribution_full} for three datasets including MSCOCO~\cite{mscoco}, A-OKVQA~\cite{a-okvqa}, and GQA~\cite{gqa} with \emph{adversarial} setting. Each dataset consists of 3000 pairs of image-questions. It can be seen that the energy score induced by the penultimate layer is  generally the lowest, and that this layer is predominantly utilized for the subsequent decoding process.

\subsection{Additional experiments}
We further evaluate our method on DeepSeek-VL-1.3B~\cite{deepseek} and LLaVA-1.5-13B with 4-bit quantization. One can see from Tab.~\ref{llava-scale} that our method consistently outperforms greedy decoding.

 \begin{table}[h!]
\scriptsize
\begin{adjustbox}{width=0.9\linewidth,center}
 \begin{tabular}{lllllllll}
  \\ [-2em]
 \toprule
  \textbf{Models}  & \textbf{Datasets}             & \textbf{Decoding}                         & Accuracy$\uparrow$ & Precision & Recall & F1 Score$\uparrow$ & $\Delta_\text{gap} \downarrow$   \\  
\midrule

\multirow{6}{*}{\shortstack{DeepSeek-VL-1.3B }}  &\multirow{2}{*}{MSCOCO}
  &Greedy  & 62.14 & 58.82 & 95.64 & 72.79 & 26.07 \\
  & &\cellcolor{mygray}Energy  &\cellcolor{mygray}\best{68.29} & \cellcolor{mygray}65.18&\cellcolor{mygray}84.37&\cellcolor{mygray}\best{73.54}&\cellcolor{mygray}\best{11.27}  \\

   &\multirow{2}{*}{A-OKVQA} 
   &  Greedy & 60.54 & 57.77 & 96.53 & 72.26 & 23.97   \\
  & &\cellcolor{mygray}Energy  &\cellcolor{mygray}\best{69.34} &\cellcolor{mygray}66.73 &\cellcolor{mygray}89.60 &\cellcolor{mygray}\best{76.24} &\cellcolor{mygray}\best{13.05}  \\
  &\multirow{2}{*}{GQA} 
   &  Greedy &59.24 & 57.17 & 94.25 & 71.16 & 22.64   \\
  & &\cellcolor{mygray}Energy  &\cellcolor{mygray}\best{65.50} &\cellcolor{mygray}64.10 &\cellcolor{mygray}85.72 &\cellcolor{mygray}\best{73.10}&\cellcolor{mygray}\best{12.68}  \\

\midrule

\multirow{6}{*}{\shortstack{LLaVA-1.5-13B \\ (4-bit quantization)}}  & \multirow{2}{*}{MSCOCO}   
& Greedy  & 85.08 & 80.69 & 93.02 & 86.31 & 7.94 \\
& &\cellcolor{mygray}Energy &\cellcolor{mygray}\best{85.30} &\cellcolor{mygray}81.48 &\cellcolor{mygray}92.09 &\cellcolor{mygray}\best{86.35} &\cellcolor{mygray}\best{6.79} \\
  & \multirow{2}{*}{A-OKVQA} 
   &  Greedy  & 75.98 & 69.17 & 95.67 & 80.14 & 19.69  \\
 & &\cellcolor{mygray}Energy  &\cellcolor{mygray}\best{76.33} &\cellcolor{mygray}69.60 &\cellcolor{mygray}95.53 &\cellcolor{mygray}\best{80.37} &\cellcolor{mygray}\best{19.20}  \\
 & \multirow{2}{*}{GQA} 
   &  Greedy  & 74.52 & 67.54 & 97.00 & 79.45 & 22.48  \\
 & & \cellcolor{mygray}Energy  &\cellcolor{mygray}\best{74.82} &\cellcolor{mygray}67.85 &\cellcolor{mygray}96.93 &\cellcolor{mygray}\best{79.64} &\cellcolor{mygray}\best{22.11}   \\

\bottomrule
\end{tabular}
\end{adjustbox}
\caption{ \footnotesize{\emph{Average performance on POPE benchmark} over three settings ( random, popular and adversarial). }}
\label{llava-scale}
\vspace{-1.5em}
\end{table}



\end{document}